\documentclass[journal]{IEEEtran}
\usepackage{amsmath,epsfig}
\usepackage{authblk}
\usepackage{amssymb}
\usepackage{url,color}
\usepackage{todonotes}
\usepackage{multirow}
\usepackage{arydshln}

\usepackage{pifont}
\newcommand{\cmark}{\ding{51}}%
\newcommand{\xmark}{\ding{55}}%

\DeclareMathOperator{\med}{median}

\definecolor{darkgreen}{rgb}{0,0.6,0.2}
\definecolor{orange}{rgb}{1.0,0.5,0.0}

\makeatletter
\let\MYcaption\@makecaption
\makeatother

\usepackage[font=footnotesize]{subcaption}

\makeatletter
\let\@makecaption\MYcaption
\makeatother

\begin{document}
%
\title{Urban Land Cover Classification with Missing Data Modalities Using Deep Convolutional Neural Networks}

\author{Michael~Kampffmeyer,~\IEEEmembership{Student Member,~IEEE,}
        Arnt-B{\o}rre~Salberg,~\IEEEmembership{Member,~IEEE,}
        and~Robert~Jenssen,~\IEEEmembership{Member,~IEEE}
\thanks{M. Kampffmeyer and R. Jenssen are with the Machine Learning Group, UiT The Arctic University of Norway, 9019 Troms{\o}, Norway, http://site.uit.no/ml/}
\thanks{A.-B. Salberg and R. Jenssen are with the Norwegian Computing Center, 0373 Oslo, Norway.}
}

\markboth{}%
{Kampffmeyer \MakeLowercase{\textit{et al.}}: Urban Land Cover Classification with Missing Data Modalities Using Deep Convolutional Neural Networks}
%

\maketitle

\begin{abstract}
Automatic urban land cover classification is a fundamental problem in remote sensing, e.g.\ for environmental monitoring. The problem is highly challenging, as classes generally have high inter-class and low intra-class variance. Techniques to improve urban land cover classification performance in remote sensing include fusion of data from different sensors with different data modalities. However, such techniques require all modalities to be available to the classifier in the decision-making process, i.e.\ at test time, as well as in training. If a data modality is missing at test time, current state-of-the-art approaches have in general no procedure available for exploiting information from these modalities. This represents a waste of potentially useful information. We propose as a remedy a convolutional neural network (CNN) architecture for urban land cover classification which is able to embed all available training modalities in a so-called hallucination network. The network will in effect replace missing data modalities in the test phase, enabling fusion capabilities even when data modalities are missing in testing. We demonstrate the method using two datasets consisting of optical and digital surface model (DSM) images. We simulate missing modalities by assuming that DSM images are missing during testing. Our method outperforms both standard CNNs trained only on optical images as well as an ensemble of two standard CNNs. We further evaluate the potential of our method to handle situations where only some DSM images are missing during testing. Overall, we show that we can clearly exploit training time information of the missing modality during testing.
\end{abstract}

\begin{IEEEkeywords}
Deep learning, convolutional neural networks, remote sensing, missing data modalities, land cover classification
\end{IEEEkeywords}

\IEEEpeerreviewmaketitle

\section{Introduction}
\label{sec:intro}
More than half of the world population now lives in cities, and 2.5 billion more people are expected to move into cities by 2050 \cite{un2015world}.
Although constituting only a small percentage of global land cover, urban areas significantly alter climate, biogeochemistry, and hydrology at local, regional, and global scales. Thus, in order to support sustainable urban development, accurate information on the urban land cover is needed.

\begin{figure}[htb]
\centering
\includegraphics[trim=0 0 0 0,clip,width=1.0\linewidth]{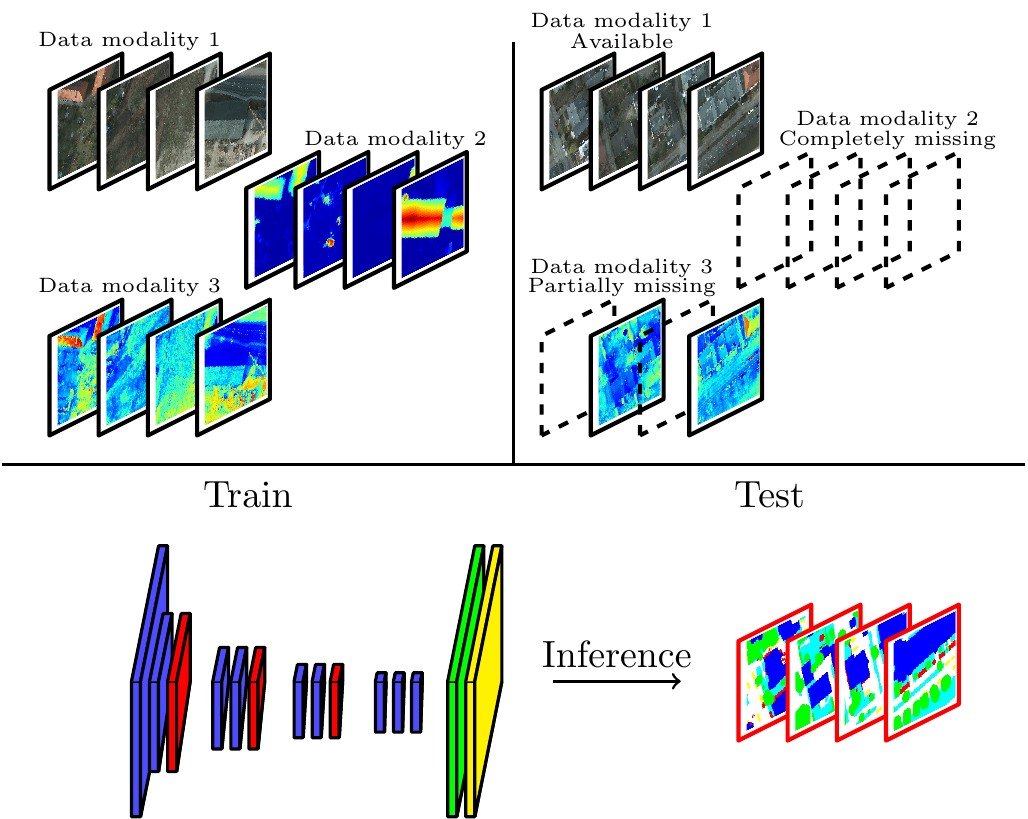}
\caption{A brief illustration of the issue addressed in this work. We propose a method to effectively produce urban land cover classification when some data modalities are missing partially or completely during the test phase. For instance, the top part of the figure illustrates a scenario, where data modality 2 is completely missing during testing and data modality 3 is missing for some of the test images. Our method leverages all available training modalities (top left part of the figure) to increase overall performance when performing inference (bottom part of figure).}
\label{fig:motivation}
\end{figure}

Due to the successes of deep learning architectures, convolutional neural networks (CNNs) have found increased use in the field of remote sensing~\cite{kampffmeyer2016semantic, maggiori2016fully, krylov2016large, volpi2017dense, audebert2017fusion}, outperforming more traditional approaches~\cite{lagrange2015benchmarking}.
The strength of CNNs is their ability to learn features that exploit the spatial context and thereby provide land cover maps with high accuracy. Most modern approaches to urban land cover classification rely on deep convolutional neural networks with \emph{multiple} modalities~\cite{kampffmeyer2016semantic, maggiori2016fully, krylov2016large, volpi2017dense, audebert2017fusion}, effectively fusing information contained in the individual data modalities. Here we use the term \emph{data modalities} to refer to data obtained by different imaging sensors, potentially operating in different frequency ranges and/or using a different imaging principle.

Data fusion is a frequently studied topic in land cover classification since it often leads to improved accuracy. It aims to integrate the information acquired potentially with different spatial resolution, spectral bands and imaging modes from sensors mounted on satellites, aircraft and ground platforms to produce fused data. This leads to fused data, which contains more detailed information than each of the individual sources \cite{zhang2010, bovolo2015}. Several approaches to fusing data modalities exist for CNNs, however, these approaches generally break down during the testing stage, when not all modalities are available. A concrete example, where all training data modalities are likely to not be present during testing is disaster monitoring and assessment systems. These systems are often time critical, especially in highly populated areas, and require analysis immediately after disasters happen. Due to this requirement, it is likely that not all data modalities have been gathered.

In this work, we propose an approach to remedy this problem. The key contribution is an approach to urban land cover mapping that exploits the relationships between the various modalities in the training dataset to improve performance when certain data modalities are missing during the test-phase. The approach further provides the flexibility to use all data modalities during testing in scenarios where they are available. We illustrate the problem that we consider in this work in Figure~\ref{fig:motivation}. Concretely we focus on how to exploit all available training modalities in the following three problem scenarios:

{\bf Problem Scenario 1}: \emph{One modality is completely missing during testing, however, since the modality is available during training, we would like to exploit the available training data.}

{\bf Problem Scenario 2}: \emph{One modality is partly missing during testing. Given a set of images, a given data modality is missing only for a subset of images. We would like to model the missing data modality for these images, but still want to make use of the data modality for the test images where it is available.}

{\bf Problem Scenario 3}: \emph{Multiple modalities are either completely or partly missing during testing. This extends problem scenario one and two to multiple missing data modalities.}

Keeping to the example of disaster monitoring and more specifically flood monitoring, these three problem scenarios can, for instance, occur in the following situations. One common approach to mapping and detection of floods is the use of Synthetic Aperture Radar (SAR)~\cite{serpico2012}, but very high-resolution optical data from airborne sensors can be incorporated to improve performance~\cite{serpico2012}. However, this data is not always available during the inference phase (Problem Scenario 1). Problem Scenario 2 allows us to use a single model to perform inference on these images, even if only parts of the area is covered by the high-resolution optical images, for instance, due to the high cost of acquiring it. Finally, when extending the task of flood mapping to flood emergency management with tasks like damage analysis~\cite{brunner2010, gueguen2015} and estimation of elements-at-risk and vulnerability~\cite{serpico2012}, it can be beneficial to map buildings and infrastructure with the help of additional data modalities, such as LIDAR (Light Detection And Ranging) based DSMs. However, adding the LIDAR-based DSMs will lead to potentially multiple missing modalities (Problem Scenario 3).

The proposed system will build upon the \emph{hallucination network} strategy proposed by Hoffman et al.~\cite{Hoffman_2016_CVPR} for training CNNs for object detection from data modalities when one of the modalities is completely missing during testing. We extend their work to land-cover classification (Problem Scenario 1) and extend the underlying idea of hallucination networks to handle situations where multiple modalities are missing (Problem Scenario 3). We further analyze its potential to handle both situations where data modalities are completely missing and situations where data modalities are only missing for some of the test images (Problem Scenario 2). Section~\ref{approach} describes the approach taken in detail. 

Note, that the aforementioned problems are inherently different from the more traditional missing data scenarios that have been investigated previously in a remote sensing context and to the author's knowledge, this is the first time that these problems have been addressed in remote sensing.

Another challenge that is often encountered when designing classifiers for land cover mapping is class imbalance. 
Land covers within the area of interest are often highly imbalanced, where some land cover types are frequent, whereas others are rare.
Moreover, many objects of interest in remote sensing are small compared to the overall image.
A solution based on optimizing the overall classification accuracy is often not satisfactory since 'small' classes will often be suppressed~\cite{estabrooks2004multiple}. In a recent paper~\cite{kampffmeyer2016semantic}, the influence of imbalanced classes was reduced by introducing weights to the cost function. Classes with few samples were given high weights, whereas classes with many samples were given small weights. 
Due to the inherent difficulties in urban land cover classification caused by small objects and imbalanced classes, we specially tailor the network by incorporating a class-wise median frequency balancing approach into the cost function.

A preliminary version of this paper appeared in~\cite{kampffmeyer2017igarss}. Here, we extend our work by 
(i) providing a more thorough literature background discussion,
(ii) extending and evaluating the methodology for scenarios where multiple data modalities are not present during the test phase (Problem Scenario 3),
(iii) expand the experiment section,
(iv) evaluate the effect of medium frequency balancing on the small (imbalanced) classes, and
(v) evaluate the potential of the trained model to also handle situations where both modalities are available during testing (Problem Scenario 2).

This paper is structured as follows. Section~\ref{sec:related} provides an overview of the background literature. Section~\ref{sec:dataset} introduces the datasets used in our work and the evaluation procedure. In Section~\ref{approach} we present the methodology and training procedure. Experimental evaluation of the method is performed in Section~\ref{sec:experiments} and finally in Section~\ref{sec:conclusion} we draw conclusions and point to future directions.

\begin{figure*}[ht]
\begin{minipage}[t]{0.49\textwidth}
    \centering
    \begin{subfigure}[b]{0.48\textwidth}
        \centering
        \includegraphics[trim=0 0 0 0,clip,width=\textwidth]{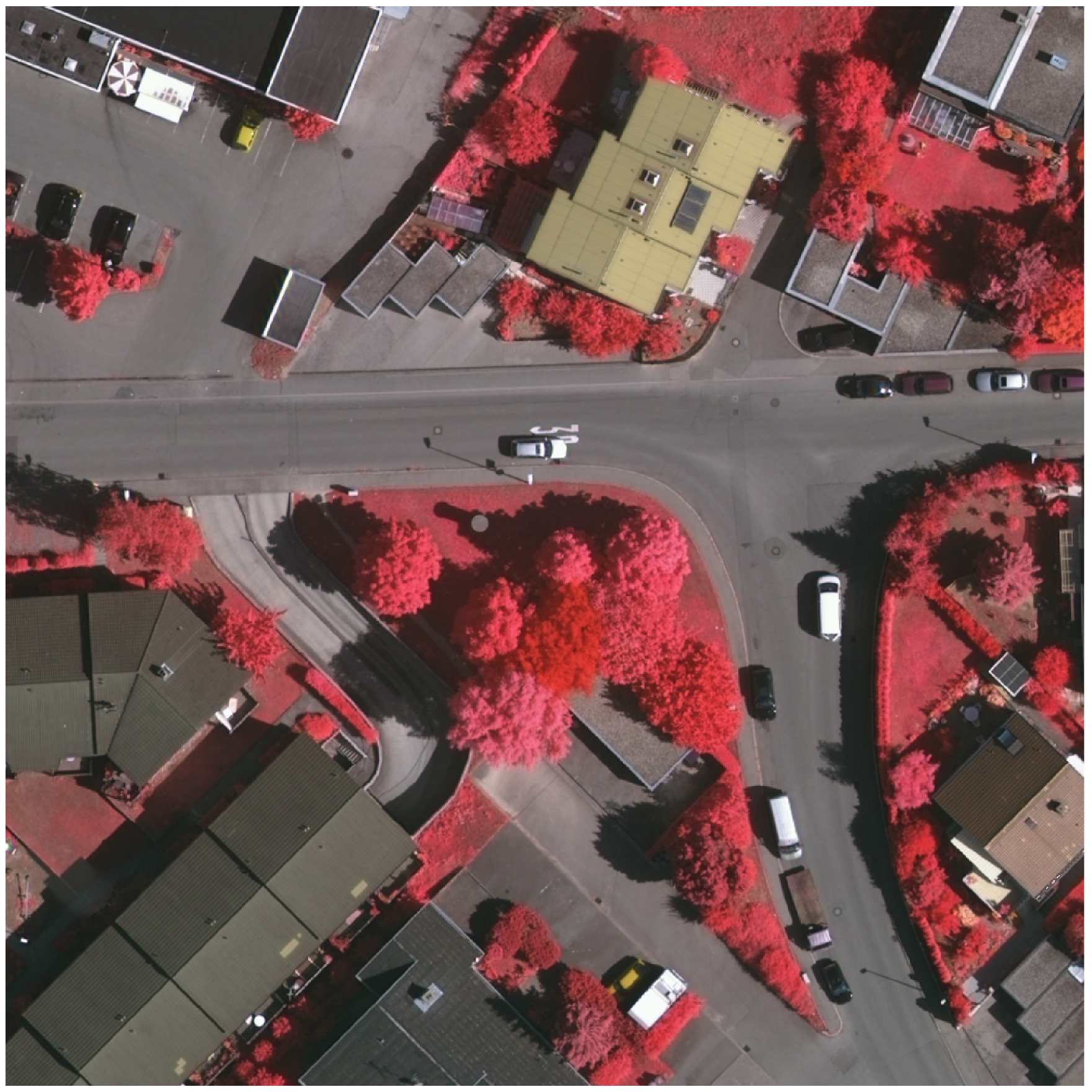}
    \end{subfigure}
    \begin{subfigure}[b]{0.48\textwidth}
        \centering
        \includegraphics[trim=0 0 0 0,clip,width=\textwidth]{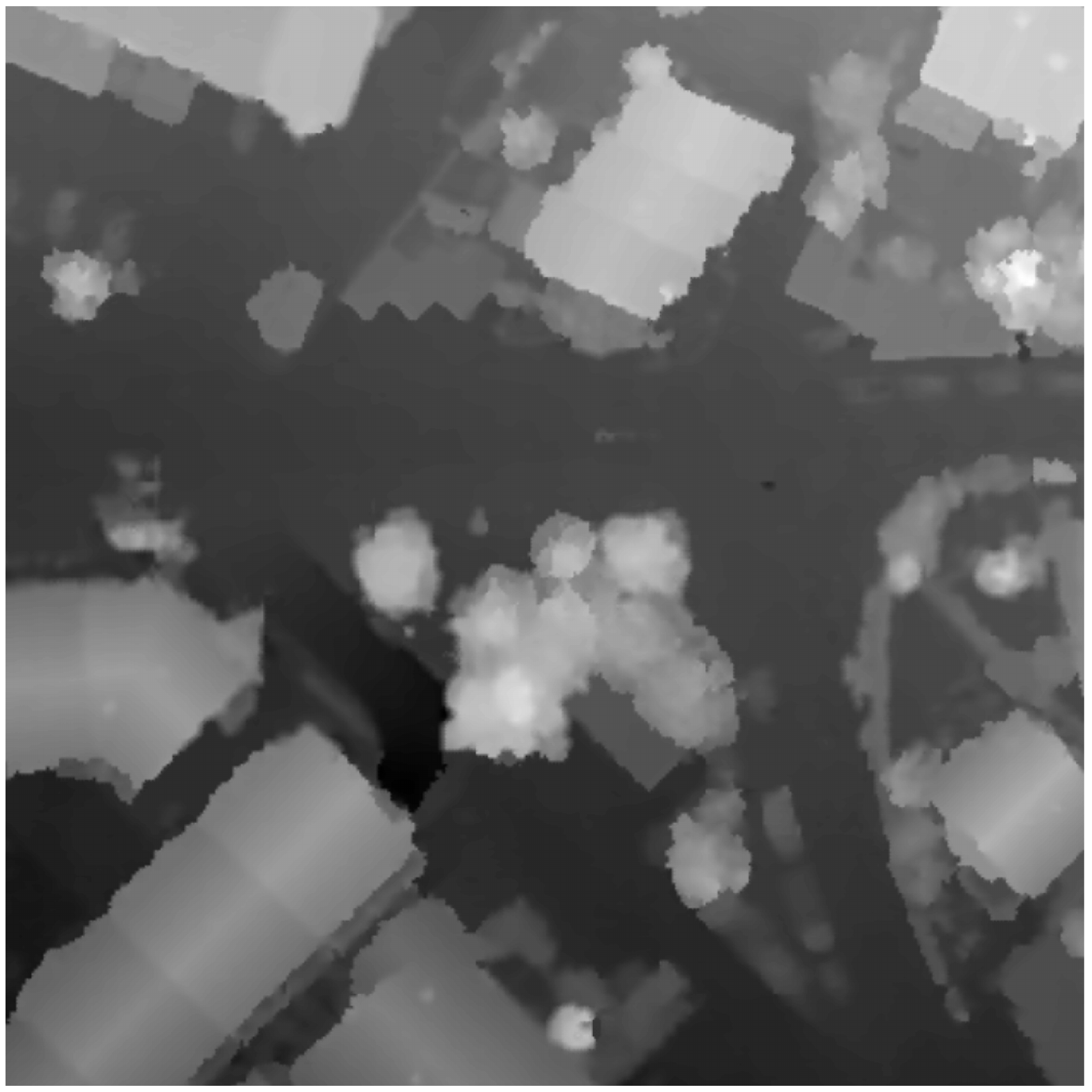}
    \end{subfigure}\\
    \begin{subfigure}[b]{0.48\textwidth}
        \centering
        \includegraphics[trim=0 0 0 0,clip,width=\textwidth]{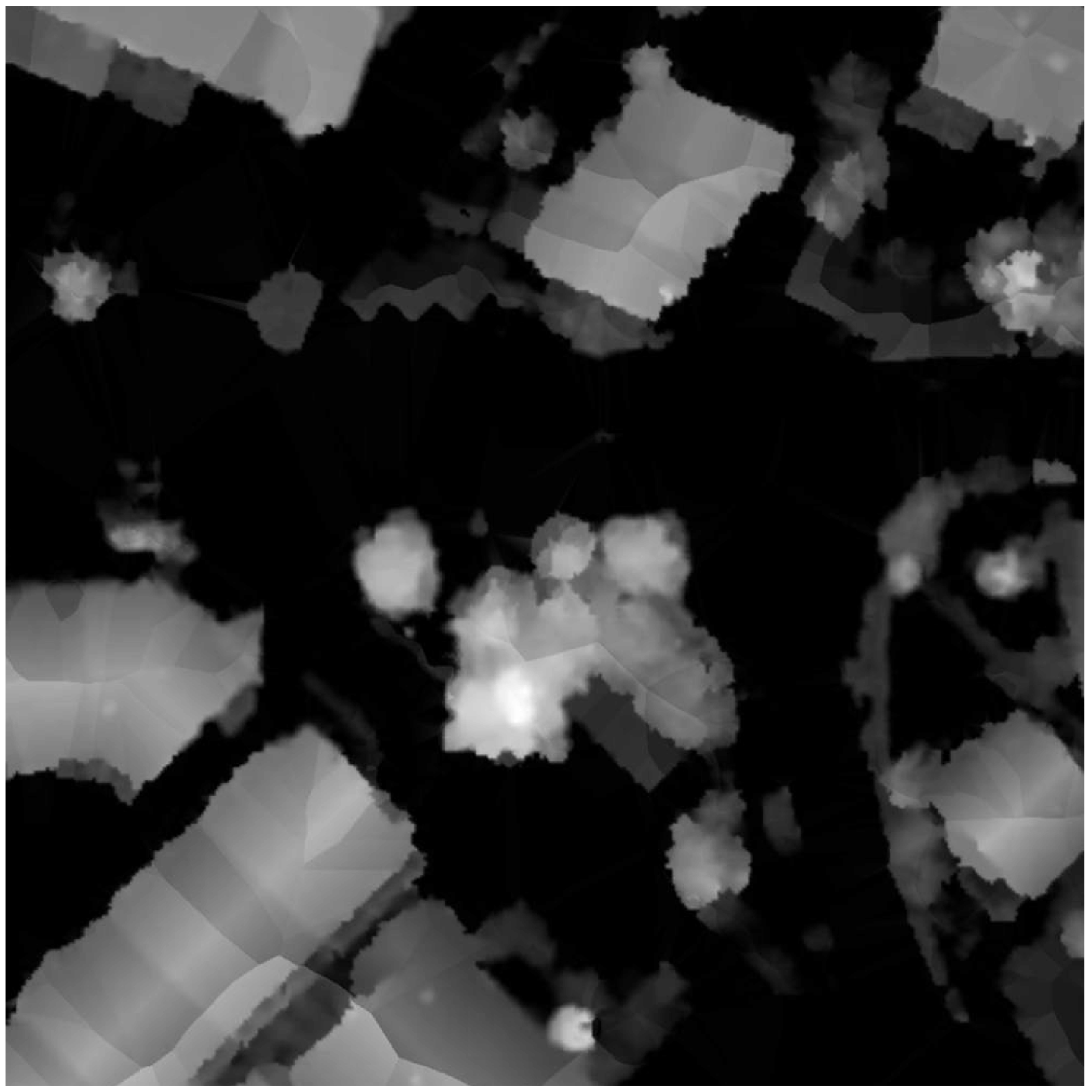}
    \end{subfigure}
    \begin{subfigure}[b]{0.48\textwidth}
        \centering
        \includegraphics[trim=0 0 0 0,clip,width=\textwidth]{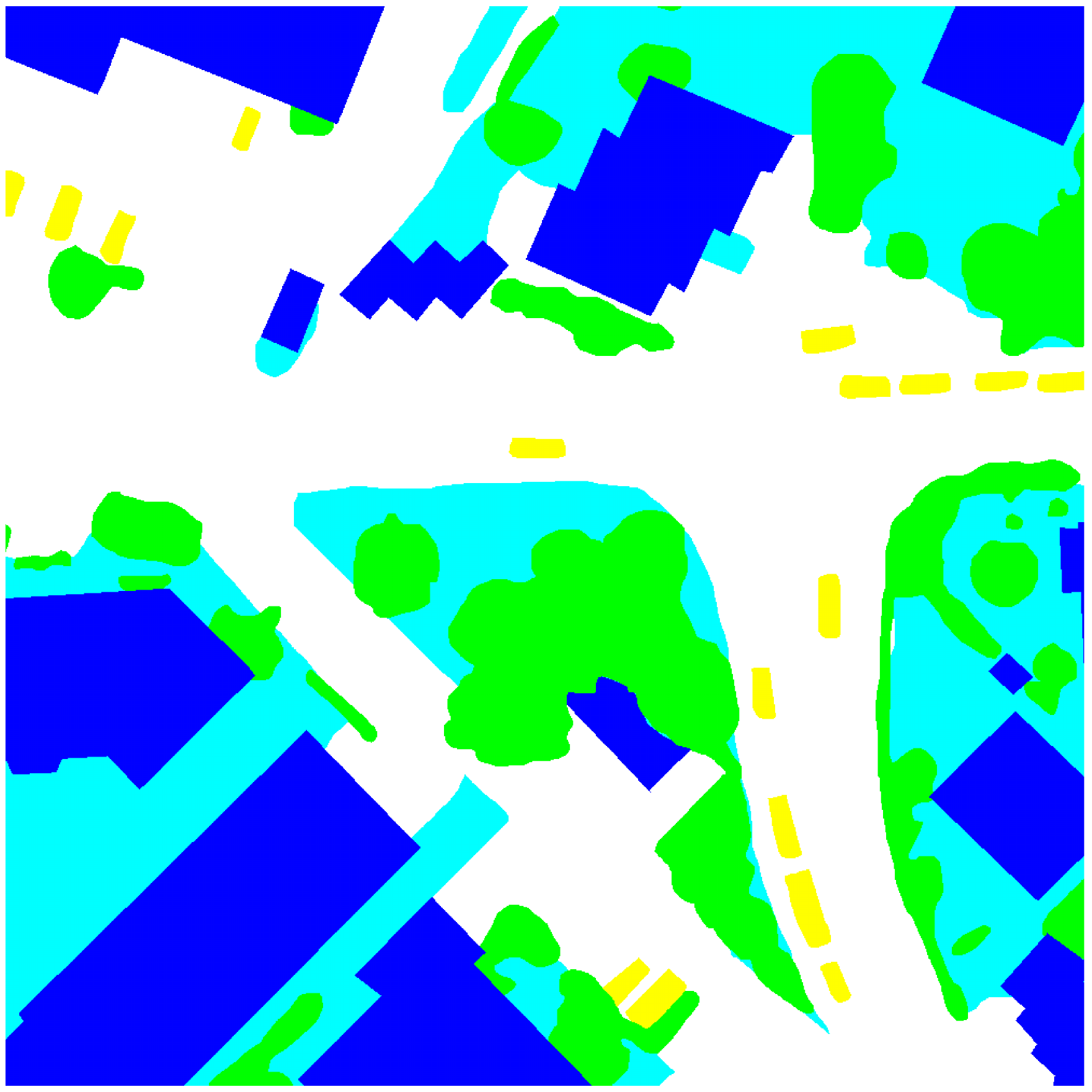}
    \end{subfigure}
    \caption{A small example patch from the Vaihingen validation dataset. From left to right and top to bottom: RG+I image, DSM, normalized DSM, and ground truth image. It illustrates the difference in size between classes, such as the car class (yellow) and the building class (blue).}
    \label{fig:exampleData}
\end{minipage}\hspace{0.2cm}
\begin{minipage}[t]{0.49\textwidth}
    \centering
    \begin{subfigure}[b]{0.48\textwidth}
        \centering
        \includegraphics[trim=0 0 0 0,clip,width=\textwidth]{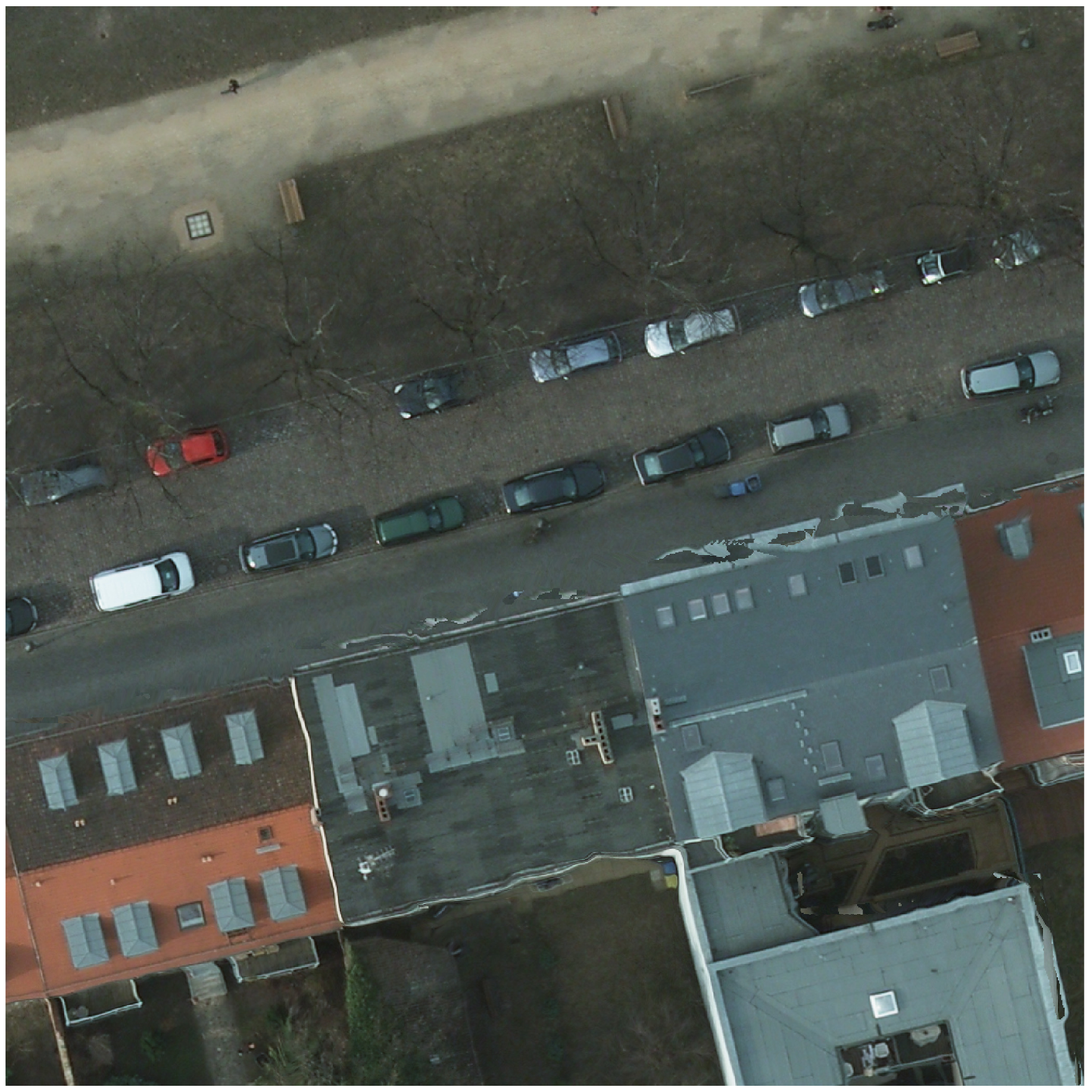}
    \end{subfigure}
    \begin{subfigure}[b]{0.48\textwidth}
        \centering
        \includegraphics[trim=0 0 0 0,clip,width=\textwidth]{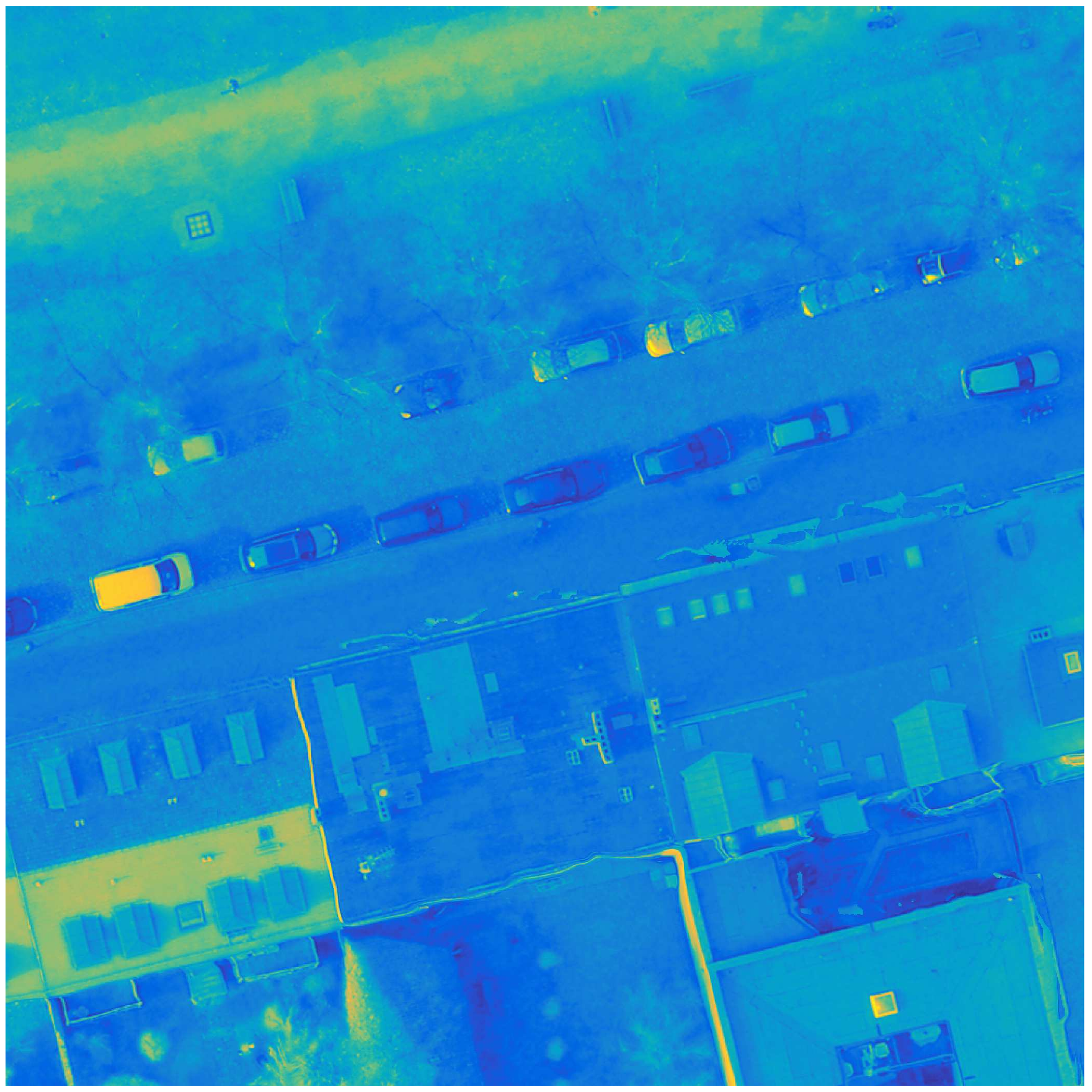}
    \end{subfigure}\\
    \begin{subfigure}[b]{0.48\textwidth}
        \centering
        \includegraphics[trim=0 0 0 0,clip,width=\textwidth]{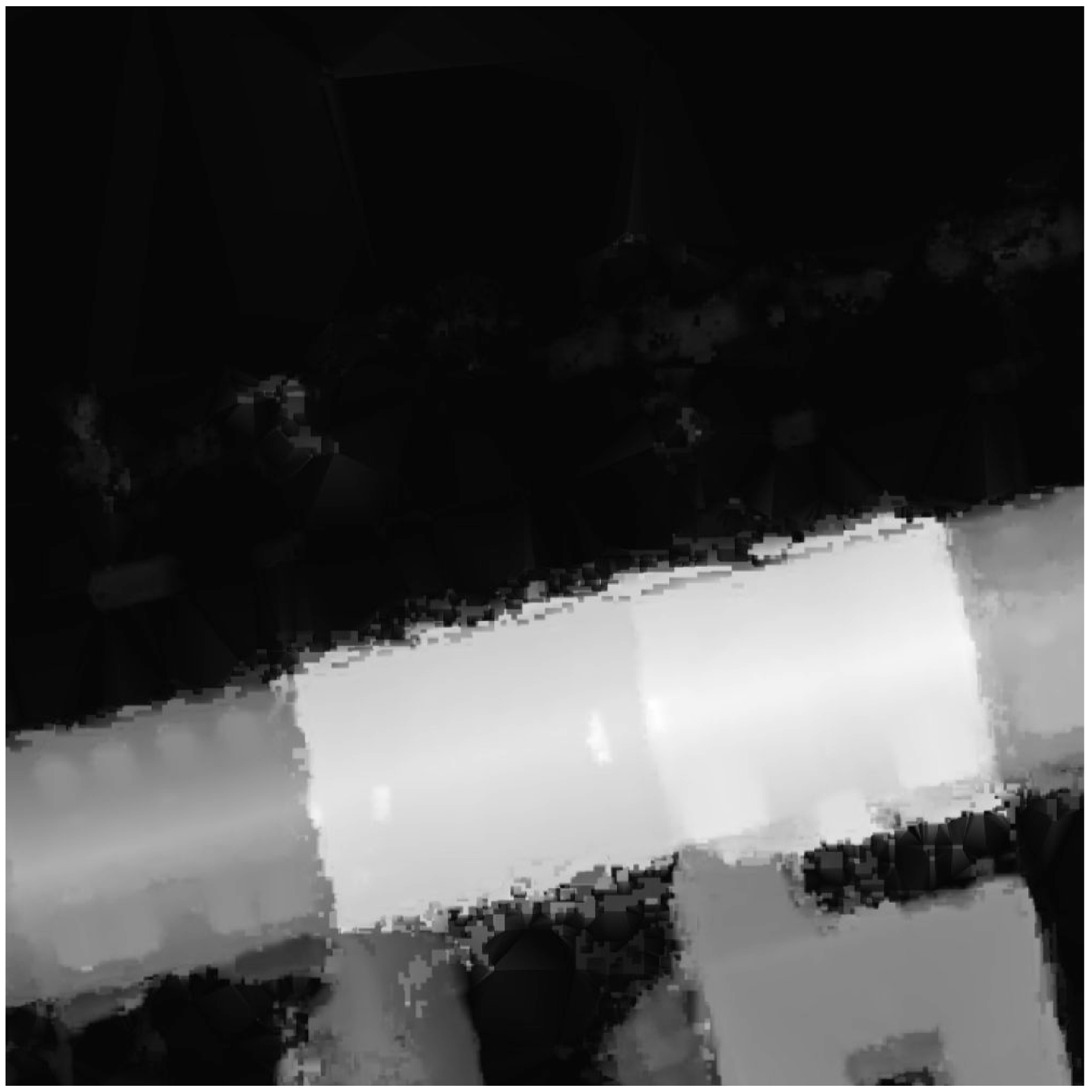}
    \end{subfigure}
    \begin{subfigure}[b]{0.48\textwidth}
        \centering
        \includegraphics[trim=0 0 0 0,clip,width=\textwidth]{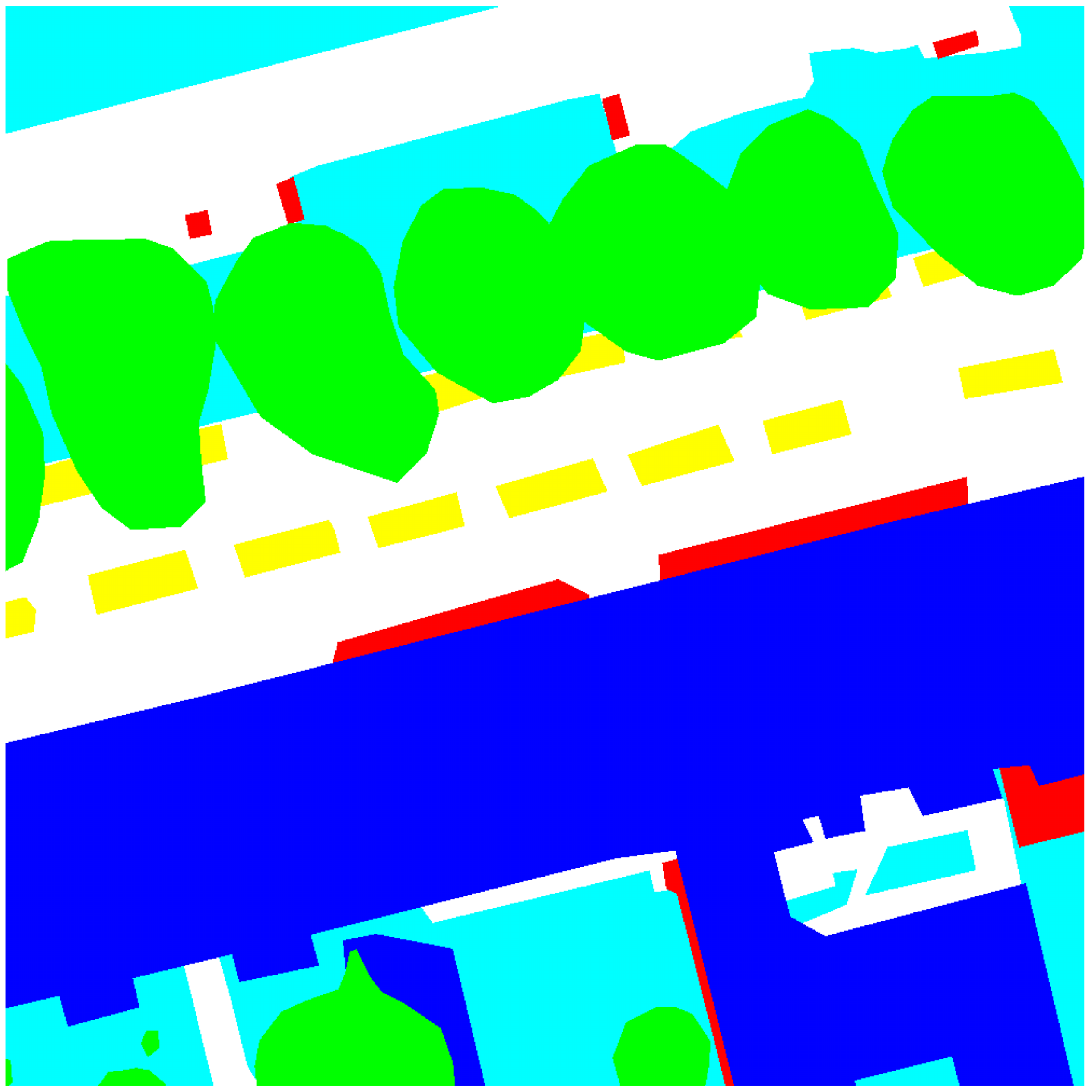}
    \end{subfigure}
    \caption{A small example patch from the Potsdam validation dataset. From left to right and top to bottom: RGB image, IR, normalized DSM, and ground truth image. The DSM modality has been omitted in this figure, as we only use the normalized DSM in our experiments.}
    \label{fig:exampleDatab}
\end{minipage}
\end{figure*}

\section{Background}
\label{sec:related}
Our approach to segmentation builds on the recent successes that deep learning techniques have achieved for image segmentation using CNNs. Instead of requiring elaborate feature design, CNNs are able to learn relevant features from data. Even though segmentation can be viewed as a pixel-wise classification problem, currently, most state-of-the-art CNN models for image segmentation are inspired by the idea of fully convolutional pixel-to-pixel end-to-end learnable architectures~\cite{long2015fully}. These architectures generally consist of an encoder-decoder architecture, where the encoder creates a lower resolution representation of the image and the decoder produces the pixel-wise prediction from this representation. Upsampling in the decoder can be performed in several ways, with one common approach being the fractional-strided convolution (also referred to as Deconvolution)~\cite{long2015fully}. Alternative approaches avoid the explicit learning of an upsampling by using the max-pooling locations from the downsampling stage to place activations, thereby achieving a sparse upsampling map, which is then converted into a dense representation with the help of convolutional filters~\cite{badrinarayanan2015segnet}. All layers in these networks are based on convolutions and do not make use of, as in previous approaches, fully connected layers, which allows their application on test images of varying size.

Lately, deep learning and CNNs have also increasingly been applied to remote sensing tasks such as for instance domain adaptation~\cite{riz2016domain}, but especially for object detection and image segmentation in remote sensing. For instance, Paisitkriangkrai et al.~\cite{paisitkriangkrai2015effective} proposed a scheme for high-resolution land cover classification using a combination of a patch-based CNN and a random forest classifier that is trained on hand-crafted features. To increase the classification accuracy further, a conditional random field (CRF) was used to smooth the final pixel labeling results. Further, Krylov et al.~\cite{krylov2016large} propose the use of a patch-based neural network to perform land cover classification in SPOT-5 imagery. More recent works, to a large extent, make use of fully convolutional architectures~\cite{kampffmeyer2016semantic,maggiori2016fully,volpi2017dense}, achieving better overall accuracy and being computationally more efficient~\cite{kampffmeyer2016semantic,volpi2017dense}. Building on the good performance of fully convolutional architectures for land cover classifications, two-stage systems to object detection have been proposed, where the segmentation is followed by a classification step~\cite{audebert2017segment} and novel approaches to fusing heterogeneous data modalities in a land cover classification context have been proposed~\cite{audebert2017fusion}.

Missing data is a common problem in real-world data that can occur due to for instance sensors being corrupt. In the traditional sense, the term missing data has been used to refer to partly missing data in a given data modality. In remote sensing missing data can also occur due to the inherent properties of certain remote sensing systems. The occurrence of cloud cover in optical images and the acquisition of images during the night using optical sensors are two examples of missing data that can occur due to the inherent nature of the sensor.
Common approaches to handling missing data for pattern recognition can be grouped into four main groups~\cite{garcia2010pattern}. Incomplete data can either be deleted, leading to the removal of potentially large amounts of training data, or missing values can be imputed~\cite{schafer1997analysis}. Alternatively, model-based approaches can be used, such as e.g. the expectation-maximization algorithm by modeling the data distribution, or approaches can be designed that incorporate missing data into the machine learning approach, such as in the case of e.g. decision trees~\cite{garcia2010pattern}.

Traditionally, most works on handling missing data in remote sensing are based on partly missing data, situations where only parts of a modality are available. One common approach of many algorithms is to simply discard these incomplete data samples~\cite{aksoy2009land}. However, more advanced methods have been proposed, such as for instance, by Latif and Mercier~\cite{latif2010self}, who leverage Self-Organizing Maps to estimate and impute missing values. Further, various data imputation techniques have been used to fill the missing parts in the data~\cite{shen2015missing}. However, imputation may often produce suboptimal results, since the assumption of missingness at random~\cite{schafer1997analysis} may not hold when missingness appears due to the properties of the sensors.
Other works that utilize the missingness directly as part of the machine learning approach include Salberg and Jenssen~\cite{salberg2012land}, who propose to train SVMs for all types of missingness patterns for landcover classification, and  Aksoy et al.~\cite{aksoy2009land}, who present a decision tree based model to data fusion when some modalities are partly missing. 

However, to the author's knowledge, none of these approaches are able to exploit the knowledge provided by all data sources to improve the accuracy when one or more data sources are completely missing. This aspect, concerning missing modalities, is the focus of this work.


\section{Benchmark datasets for urban land cover classification}
\label{sec:dataset}
To perform urban land cover classification we require large-scale datasets. In this work, we focus on two commonly used benchmark datasets that each consist of more than one data modality. The datasets are the ISPRS Vaihingen and the ISPRS Potsdam 2D semantic labeling benchmark datasets~\cite{isprsdataset}. The Vaihingen dataset consists of $33$ high-resolution true ortho photo images with the three bands corresponding to near infrared, red and green bands (RG+I).
The images are acquired with a ground sampling distance of $9$ cm and of varying size (ranging from approximately 3 million to 10 million pixels). Ground truth is available for $16$ images. The Potsdam dataset consists of $38$ four-channel true ortho photo images with the channels corresponding to red, green, blue and infrared (RGB+I). The ground sampling distance is $5$ cm with each image containing 36 million pixels and with ground truth available for $24$ images. Additionally, both the digital surface model (DSM) and the normalized DSM (produced by Gerke~\cite{markususe}) is available for all images in both datasets. Figures~\ref{fig:exampleData} and~\ref{fig:exampleDatab} show two example patches, one from the Vaihingen dataset and one from the Potsdam dataset.
The datasets contain six classes. Impervious surface (white in ground truth), buildings (blue), low vegetation (cyan), trees (green), cars (yellow) and background/clutter (red). To evaluate the approach and its use for missing data modalities for urban land cover classification, the normalized DSM was included only during the training phase. 

In this paper, we will follow the guidelines and the evaluation metrics that have been specified by the ISPRS~\cite{isprsdataset} to evaluate performance on these datasets. The evaluation metrics are the F1-score and accuracy. The F1-score is computed per class and overall as
\begin{align}
    \text{F1}=2\cdot \text{precision}\cdot \text{recall}/(\text{precision}+\text{recall}) \; .
\end{align}
The accuracy is the percentage of correctly labeled pixels measured as $\sum_i n_{ii} / \sum_i t_i$, where $n_{ij}$ denotes the number of pixels of class $i$ that are classified as class $j$ and $t_i$ the total number of pixels in class $i$. We report the overall class-independent and also the mean class accuracy. 

Following the ISPRS specifications, the class boundaries are eroded with a disk of radius 3 and ignored in the evaluation to reduce boundary effects.
For evaluation, the labeled part of the Vaihingen dataset is divided into a training set, a validation set and a test set containing $11$, $2$ and $3$ images, respectively. The Potsdam dataset is similarly split into training, validation and test set ($19$, $2$ and $3$ images). The models are trained on the training dataset, hyperparameters are tuned according to the validation dataset and the final accuracy is reported on the test dataset.

\section{Approach}
\label{approach}
In this section, we describe the implementation and the training details of the proposed architecture. We start by explaining the main idea of hallucination networks in~\ref{sec:hal} since this framework is at the core of the proposed approach. The concept of median frequency balancing for handling small classes is explained in~\ref{sec:mfb}. In~\ref{sec:obj_seg} we provide details on the overall implementation and describe how we combine median frequency balancing and the idea of hallucination network to perform urban land cover classification in scenarios, where a data modality is missing during the inference phase. Finally, we propose in~\ref{sec:multiple_missing} an extension of the hallucination network methodology from dealing solely with one missing modality to handling the case of multiple missing modalities.

Our proposed method builds on a common approach to data fusion using convolutional neural networks, where features are extracted independently from each data modality using separate dedicated CNNs and the final scores are combined. Such approaches have previously shown promising results for combining multiple data modalities, such as for example RGB and Depth information~\cite{gupta2014learning, audebert2017fusion}.

\subsection{Hallucination Networks}
\label{sec:hal}
Hallucination Networks~\cite{Hoffman_2016_CVPR} are recent attempts to use data modalities that are available solely during the training phase in test time to improve object detection performance. They are inspired by recent works on transferring information between neural networks through network distillation, which initially was proposed as a way to compress the knowledge of large models into a smaller one by training small neural networks on the softmax output of the larger network~\cite{hinton2015distilling}. 

Utilizing all available training data modalities in the case of missing data modalities during the inference phase is done by adding an additional network, the hallucination network, in addition to networks for the available modalities. This network takes a data modality as input that we assume to be available during both training and testing and attempts to learn a mapping function from this modality to the modality that is missing during testing. This is done by adding a loss for the mid-level features of the hallucination network to mimic the mid-level features of the data modality that is missing during inference. In the case that the modality is missing during testing, inference is performed by not only passing the available modality to its dedicated network but in addition also passing it to the hallucination network. The raw scores of the two networks are then averaged and the land-cover classification map is produced through a final softmax layer. 

\begin{figure}[tbp]
\begin{minipage}[b]{1.0\linewidth}
  \centering
    \includegraphics[trim=0 0 0 0,clip,width=1.0\linewidth]{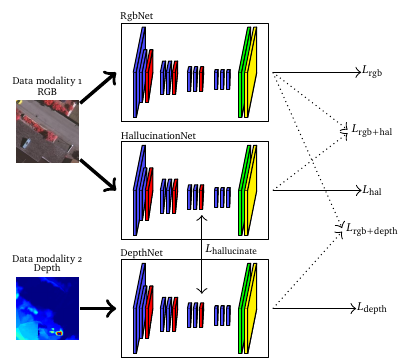}
\end{minipage}
\caption{The network architecture that addresses Problem Scenario 1 of the proposed method. Note that we choose the layer where the hallucination loss is applied ($D$) to be the third pooling layer based on the fact that Hoffman et al.~\cite{Hoffman_2016_CVPR} reported the best accuracy when hallucination is applied on the mid/late-level features. A more thorough analysis of this choice is left for future work.}
\label{fig:Architecture}
\end{figure}

\subsection{Medium frequency balancing}
\label{sec:mfb}
Medium frequency balancing is a weighted cross-entropy loss function that has been shown to yield good performance for imbalanced classes~\cite{eigen2015predicting, badrinarayanan2015segnet, kampffmeyer2016semantic}. Instead of optimizing the commonly used cross-entropy loss, where small classes contribute less to the overall loss and are therefore often neglected compared to the larger classes, each class in the loss function is weighted by the ratio of the median class frequency and the class frequency (computed over the training dataset). The updated cost formulation is
\begin{align}
    L &= -\frac{1}{N} \sum_{n=1}^N \sum_{c\in\mathcal{C}} l_{c}^{n}\log\left(\widehat{p}_{\,c}^{\,n}\right) w_{c} \; ,
\end{align}  
where
\begin{equation}
w_{c} = \frac{\med\left(\{f_c\;  | \; c \;\in\; \mathcal{C}\}\right)}{f_c}
\end{equation}
denotes the weight for class $c$, $f_c$ is the frequency of pixels in class $c$, and $\widehat{p}_{\,c}^{\,n}$ is the softmax probability of sample $n$ being in class $c$. $l_{c}^{n}$ corresponds to the label of sample $n$ for class $c$ when the label is given in one-hot encoding, $\mathcal{C}$ is the set of all classes and $N$ is the number of samples in the mini-batch.

\subsection{Object segmentation for imbalanced classes with one missing modality}
\label{sec:obj_seg}
Our architecture for the individual networks follows the architecture of Kampffmeyer et al.~\cite{kampffmeyer2016semantic} and is based on the fully convolutional network architecture~\cite{long2015fully}. It consists of four sets of two $3\times3$ convolutional layers, each set followed by a $2\times2$ max-pooling layer and each individual convolution layer followed by a ReLU nonlinearity and batch normalization layer~\cite{ioffe2015batch}. The first convolutional layer has stride $2$ due to memory restrictions during the test-phase when considering the large images, whereas all other convolutions are of stride $1$. Three of these networks are trained jointly. The first one for the RG+I or RGB+I images, which will be referred to as RGB network for simplicity, the second one for the depth image and the third one as the hallucination network. Figure \ref{fig:Architecture} illustrates the complete Architecture that is used to address Problem Scenario 1. Training is performed on image patches of size $256\times256$, which are extracted from the original images with $50\%$ overlap and are flipped and rotated at 90-degree intervals as part of the data augmentation step. For the Potsdam dataset, we only rotate the images due to the fact that much more data is available, making additional data augmentation unnecessary.

The total loss consists of six individual losses and is
\begin{align}
\label{lossfunction}
    \begin{split}
    L &= \gamma L_{\text{hallucinate}} + L_{\text{depth}} + L_{\text{rgb}} + L_{\text{hal}}\\
    & + L_{\text{rgb+depth}} + L_{\text{rgb+hal}} \; .
\end{split}
\end{align}  
The loss is optimized using backpropagation. $L_{\text{hallucinate}}$, $L_{\text{rgb}}$, $L_{\text{depth}}$ and $L_{\text{hal}}$ are the hallucination loss, the loss of the RGB network, the loss of the depth network and the loss of the hallucination network, respectively, and $L_{\text{rgb+depth}}$ and $L_{\text{rgb+hal}}$ are the joint losses. The parameter $\gamma$ is the weight parameter for the hallucination loss. The individual losses for the three networks, $L_{\text{rgb}}$, $L_{\text{depth}}$ and $L_{\text{hal}}$, ensure that the features learned by the three individual networks will be useful for the land cover classification task also independently from each other. The joint losses, $L_{\text{rgb+depth}}$ and $L_{\text{rgb+hal}}$, instead ensure that the network performs well on all the combination of modalities that can be observed during testing. Here we assume that the data for the RGB network is always available, but that the depth data modality might be missing during testing. In the case that the depth data is missing during testing, the combination of the RGB and the hallucination network can be used for inference. On the other hand, in the case where the depth data modality is available during testing, the combination of the RGB and the depth network can be used. The potential of using the same architecture for both scenarios (Problem Scenario 2) is analyzed in Section~\ref{sec:experiments}. 

\begin{figure}[tbp]
\begin{minipage}[b]{1.0\linewidth}
  \centering
    \includegraphics[trim=0 0 0 0,clip,width=1.0\linewidth]{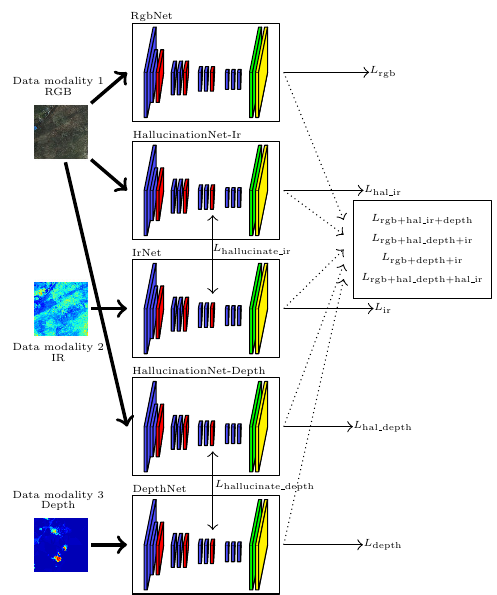}
\end{minipage}
\caption{Network architecture of the proposed method when handling multiple missing modalities (Problem Scenario 3).}
\label{fig:Architecture_multi}
\end{figure}

Following Hoffman et al.~\cite{Hoffman_2016_CVPR}, the hallucination loss is
\begin{align}
    L_{\text{hallucinate}} = ||\sigma(A_{\text{depth}}) - \sigma(A_{\text{hal}})||_2^2 \; ,
\end{align}  
where $\sigma(\cdot)$ is the sigmoid function and $A(\cdot)$ refers to the activation of the networks at a certain network depth $D$. To avoid that the depth features are adapting during the end-to-end training procedure, the learning rate for the layers before network depth $D$ are set to zero for the depth network. The hallucination loss in our experiments is based on the activations after the third pooling layer ($D$) and medium frequency balancing is used for all terms in the loss function except the hallucination loss. 
    
\paragraph*{Training details}
Training is performed by first training the RGB and depth network separately and then fine-tuning the whole architecture end-to-end. The depth network was used to initialize the hallucination network and the weight of the hallucination network loss. $\gamma$ was set such that the hallucination loss is roughly $10$ times the loss of the largest loss of the remaining terms in Eq. \ref{lossfunction}, following the example of Hoffman et al.~\cite{Hoffman_2016_CVPR}. A more thorough analysis of how to optimally set the $\gamma$ parameter is left for future work.

To avoid large variations in the magnitude of the gradients, gradient clipping~\cite{pascanu2012understanding} is performed to clip outlier gradients to an acceptable range, which is determined by monitoring the gradients during training. Training is performed using Adam~\cite{kingma2014adam}.

\subsection{Multiple missing modalities}
\label{sec:multiple_missing}
In this section, we propose an extension of the hallucination network to handle more than one missing data modality (Problem scenario 3). We focus especially on the Potsdam dataset as it comes with $3$ unique modalities, RGB, infrared and depth. Figure~\ref{fig:Architecture_multi} illustrates the updated architecture, where an additional hallucination network is introduced to handle a missing infrared data modality. The updated cost function consists of $11$ terms, which are optimized jointly following the same procedure as in the one-missing modality case, and is
\begin{align}
\label{lossfunction_multi}
    \begin{split}
    L &= \gamma (L_{\text{hallucinate\_ir}} + L_{\text{hallucinate\_depth}}) + L_{\text{ir}} + L_{\text{depth}} \\
    &+ L_{\text{rgb}} + L_{\text{hal\_depth}} + L_{\text{hal\_ir}} + L_{\text{rgb+hal\_ir+depth}} \\
    &+ L_{\text{rgb+ir+hal\_depth}} + L_{\text{rgb+ir+depth}} + L_{\text{rgb+hal\_ir+hal\_depth}} \;.
\end{split}
\end{align}  
In addition to the two hallucination losses, $L_{\text{hallucinate\_ir}}$ and $L_{\text{hallucinate\_depth}}$, which are used to transfer information from the training phase about the relationship between RGB and Depth images to the test stage, we have $5$ losses, one for each network. As in the single missing modality scenario, these losses ensure that the individual models learn good individual feature representation that can then be used for the final prediction task. In addition, losses are added for the various combinations of available and missing modalities, namely, $L_{\text{rgb+hal\_ir+depth}}$, $L_{\text{rgb+ir+hal\_depth}}$, $L_{\text{rgb+ir+depth}}$ and $L_{\text{rgb+hal\_ir+hal\_depth}}$, where the subscript indicates the network losses that are combined. These losses ensure, that the network is able to utilize the availability of data modalities during testing for the cases where modalities are only sometimes missing. For instance, given that both infrared and depth information is missing, testing is performed by using the RGB and the two hallucination networks, which means that the loss $L_{\text{rgb+hal\_ir+hal\_depth}}$ should be low. On the other hand, if for example the depth information is missing, but infrared is available, we would in the test phase use the RGB network, the depth hallucination network and the infrared network. In this case the loss $L_{\text{rgb+ir+hal\_depth}}$ should be low.

\begin{table*}[t]
\begin{center}
\begin{tabular}{lc|c|c|c|c|c|c|c|c|c|c|c|c|c|c|}
\cline{3-15}
 & & \multicolumn{2}{|c|}{Imp Surf} & \multicolumn{2}{c|}{Building} & \multicolumn{2}{c|}{Low veg} & \multicolumn{2}{c|}{Tree} & \multicolumn{2}{c|}{Car} & \multicolumn{3}{c|}{Overall}\\
\cline{1-15}
\multicolumn{1}{|c|}{Method} & \multicolumn{1}{|c|}{MFB} & F1 & Acc & F1 & Acc & F1 & Acc & F1 & Acc & F1 & Acc & Avg F1 & Avg Acc & Acc\\
\cline{1-15} 
\multicolumn{1}{|c|}{RG+I} & \cmark & 84.95 & 87.26 & 89.24 & 86.43 & 75.47 & 72.27 & 82.90 & 86.95 & 75.37 & 83.82 & 81.59 & 83.35 & 83.50\\
\cline{2-15}
\multicolumn{1}{|c|}{RG+I-ensemble} & \cmark & 85.30 & 87.83 & 89.32 & 86.14 & 76.12 & 73.52 & 83.28 & 87.01 & 78.15 & {\bf85.19} & 82.44 & 83.94 & 83.86 \\
\cline{2-15}
\multicolumn{1}{|c|}{Hallucination} & \cmark & {\bf87.15} & {\bf89.58} & {\bf91.13} & {\bf88.30} & {\bf77.05} & {\bf74.33} & {\bf83.72} & {\bf87.47} & {\bf80.92} & 84.30 & {\bf83.99} & {\bf84.80} & {\bf85.20} \\
\hdashline
\multicolumn{1}{|c|}{RG+I\&Depth} & \cmark & 88.89 & 89.33 & 93.29 & 91.42 & 77.29 & 75.68 & 83.88 & 87.38 & 81.82 & 83.42 & 85.03 & 85.45 & 86.33 \\
\hline
\hline
\multicolumn{1}{|c|}{RG+I} & \xmark & 86.83 & 89.13 & 91.49 & 89.81 & 75.30 & 75.18 & 85.67 & 86.36 & 59.00 & 44.38 & 79.66 & 76.97 & 84.96\\
\hline
\multicolumn{1}{|c|}{RG+I-ensemble} & \xmark & 87.14 & 89.47 & 91.65 & 89.79 & 75.71 & {\bf75.74} & 85.76 & 86.39 & 60.01 & 45.09 & 80.05 & 77.30 & 85.17 \\
\hline
\multicolumn{1}{|c|}{Hallucination} & \xmark & {\bf88.19} & {\bf91.14} & {\bf92.49} & {\bf90.58} & {\bf76.54} & 74.31 & {\bf86.49} & {\bf88.34} & {\bf74.75} & {\bf62.90} & {\bf83.69} & {\bf81.45} & {\bf86.22}\\
\hdashline
\multicolumn{1}{|c|}{RG+I\&Depth} & \xmark & 88.17 & 91.35 & 93.06 & 91.36 & 76.92 & 75.69 & 86.69 & 87.54 & 65.31 & 50.01 & 82.03 & 79.19 & 86.39\\
\hline
\end{tabular}
\vspace{0.2cm}
\caption{Performance of the different models for the {\bf\scshape Vaihingen} dataset. The F1 scores and accuracies are shown as percentages. Bold numbers indicate the best accuracy among the first three models. The final model $\text{RG+I\&Depth}$ is used as a reference to illustrate the overall accuracy that could be achieved by a model if all data modalities are available and no hallucination network is employed.}
\label{tab:resultsMFBVaihingen}
\end{center}
\end{table*}

\begin{table*}[t]
\begin{center}
\begin{tabular}{lc|c|c|c|c|c|c|c|c|c|c|c|c|c|c|}
\cline{3-15}
& & \multicolumn{2}{|c|}{Imp Surf} & \multicolumn{2}{c|}{Building} & \multicolumn{2}{c|}{Low veg} & \multicolumn{2}{c|}{Tree} & \multicolumn{2}{c|}{Car} & \multicolumn{3}{c|}{Overall}\\
\cline{1-15}
\multicolumn{1}{|c|}{Method} & \multicolumn{1}{|c|}{MFB} & F1 & Acc & F1 & Acc & F1 & Acc & F1 & Acc & F1 & Acc & Avg F1 & Avg Acc & Acc\\
\cline{1-15} 
\multicolumn{1}{|c|}{RGB+I} & \cmark & 86.36 & 90.57 & 92.74 & 91.84 & 81.69 & 82.73 & 83.94 & 84.78 & 88.51 & 97.61 & 86.65 & 89.51 & 85.74\\
\cline{2-15}
\multicolumn{1}{|c|}{RGB+I-ensemble} & \cmark & 87.03 & 91.28 & 93.64 & {\bf92.99} & {\bf82.19} & 82.98 & 84.49 & {\bf85.08} & {\bf89.15} & 97.86 & {\bf87.30} & 90.04 & 86.43 \\
\cline{2-15}
\multicolumn{1}{|c|}{Hallucination} & \cmark & {\bf87.26} & {\bf93.01} & {\bf93.84} & 91.41 & 82.08 & {\bf83.72} & {\bf84.81} & 84.25 & 88.17 & {\bf97.88} & 87.23 & {\bf90.05} & {\bf86.53} \\
\hdashline
\multicolumn{1}{|c|}{RGB+I\&Depth} & \cmark & 88.76 & 91.00 & 96.26 & 97.79 & 81.38 & 79.36 & 83.48 & 87.22 & 84.98 & 98.02 & 86.97 & 90.68 & 87.58 \\
\hline
\hline
\multicolumn{1}{|c|}{RGB+I} & \xmark & 86.65 & 92.82 & 93.49 & 93.07 & 81.41 & 83.20 & 81.93 & 78.34 & 88.11 & 81.78 & 86.32 & 85.84 & 85.79\\
\hline
\multicolumn{1}{|c|}{RGB+I-ensemble} & \xmark & 87.06 & 91.98 & 93.98 & {\bf95.14} & 82.22 & {\bf84.62} & 82.71 & 77.96 & 89.61 & 84.58 & 87.11 & 86.86 & 86.39 \\
\hline
\multicolumn{1}{|c|}{Hallucination} & \xmark & {\bf87.47} & {\bf94.51} & {\bf94.62} & 93.47 & {\bf82.76} & 83.13 & {\bf83.86} & {\bf80.42} & {\bf91.04} & {\bf90.62} & {\bf87.95} & {\bf88.43} & {\bf86.96}\\
\hdashline
\multicolumn{1}{|c|}{RGB+I\&Depth} & \xmark & 88.75 & 94.50 & 96.74 & 97.26 & 82.10 & 83.90 & 83.14 & 79.85 & 82.49 & 71.37 & 86.65 & 85.38 & 87.76\\
\hline
\end{tabular}
\vspace{0.2cm}
\caption{Performance of the different models for the {\bf\scshape Potsdam} dataset. The F1 scores and accuracies are shown as percentages. Bold numbers indicate the best accuracy among the first three models. The final model, $\text{RG+I\&Depth}$, is used as a reference to illustrate the overall accuracy that could be achieved by a model if all data modalities are available and no hallucination network is employed.}
\label{tab:resultsMFBPotsdam}
\end{center}
\end{table*}

\begin{table*}[t]
\begin{center}
\begin{tabular}{lc|c|c|c|c|c|c|c|c|c|c|c|c|c|c|c|}
\cline{3-15}
& & \multicolumn{2}{|c|}{Imp Surf} & \multicolumn{2}{c|}{Building} & \multicolumn{2}{c|}{Low veg} & \multicolumn{2}{c|}{Tree} & \multicolumn{2}{c|}{Car} & \multicolumn{3}{c|}{Overall}\\
\cline{1-15}
\multicolumn{1}{|c|}{Method} & \multicolumn{1}{|c|}{MFB} & F1 & Acc & F1 & Acc & F1 & Acc & F1 & Acc & F1 & Acc & Avg F1 & Avg Acc & Acc\\
\hline\hline
\multicolumn{1}{|c|}{Hallucination} & \cmark & 87.26 & {\bf93.01} & 93.84 & 91.41 & {\bf82.08} & {\bf83.72} & {\bf84.81} & {\bf84.25} & 88.17 & {\bf97.88} & 87.23 & 90.05 & 86.53\\
\hline
\multicolumn{1}{|c|}{RGB+I\&Depth (Hal)} & \cmark & {\bf87.90} & 93.00 & {\bf95.20} & {\bf94.14} & 82.01 & 82.78 & 84.52 & 83.91 & {\bf88.76} & 97.64 & {\bf87.68} & {\bf90.29} & {\bf87.20} \\
\hline
\hline
\multicolumn{1}{|c|}{Hallucination} & \xmark & 87.47 & 94.51 & 94.62 & 93.47 & {\bf82.76} & 83.13 & {\bf83.86} & {\bf80.42} & {\bf91.04} & {\bf90.62} & {\bf87.95} & {\bf88.43} & 86.96\\
\hline
\multicolumn{1}{|c|}{RGB+I\&Depth (Hal)} & \xmark & {\bf87.78} & {\bf94.92} & {\bf95.59} & {\bf95.06} & 82.65 & {\bf83.86} & 82.78 & 77.45 & 89.13 & 85.01 & 87.59 & 87.26 & {\bf87.20} \\
\hline
\end{tabular}
\vspace{0.2cm}
\caption{Performance of the hallucination {\bf\scshape Potsdam} model. The F1 scores and accuracies are shown as percentages. Bold numbers indicate the best accuracy among the two models. The final model, $\text{RG+I\&Depth (Hal)}$, corresponds to the trained hallucination model. However, in this case we do not use the hallucination network during testing, but instead assume that the depth information is available such that we can use the trained $\text{RG+I}$ and $\text{Depth}$ networks.}
\label{tab:resultsHallucinationPotsdam}
\end{center}
\end{table*}

\section{Experiments and results}
\label{sec:experiments}
In this section, we perform experiments on the Vaihingen (Section~\ref{sec:vaihingen}) and Potsdam (Section~\ref{sec:potsdam}) datasets and report both qualitative and quantitative results to address Problem Scenario 1. 
We further analyze for the trained Potsdam hallucination network its ability to perform inference in situations where suddenly both modalities are available. This is crucial to address Problem Scenario 2, the problem of partially missing data modalities, as the single trained model should not only be able to perform well in scenarios where data modalities are missing. Instead, it should still be able to exploit the information of a given data modality when it is present during testing to improve performance. 
Section~\ref{sec:missing_experiments} addresses Problem Scenario 3 and focuses on the case of multiple missing modalities for the Potsdam scenario, where both infrared and depth measurements are missing. For all datasets, we compare our proposed approach to a CNN trained only on the RGB+I (or RG+I) image. To ensure fairness with respect to the number of network parameters, we also compare the approach to an ensemble of two CNNs trained on the RGB+I (or RG+I) images. For the ensemble, the softmax output of the two CNNs was averaged during the test-phase. 

\subsection{Vaihingen}
Initially, we focus on Problem Scenario 1, the case of one missing data modality.
Table \ref{tab:resultsMFBVaihingen} shows the results for the models when trained on the Vaihingen dataset with and without medium frequency balancing. It illustrates that the hallucination network outperforms both the single RG+I model as well as the ensemble when considering overall accuracy. Increases in accuracy can be observed in most classes with large increases being observed for the impervious surface class and the building class. Given that these two classes appear similar in many of the RGB images, it is not surprising that additional depth information would benefit these classes considerably. This indicates that some of the additional information contained in the available training depth data is benefiting the test-phase of our model even though the depth data is missing during testing.

\label{sec:vaihingen}
\begin{figure}[htbp]
\begin{minipage}[b]{0.48\linewidth}
  \centering
 \includegraphics[width=1.0\linewidth]{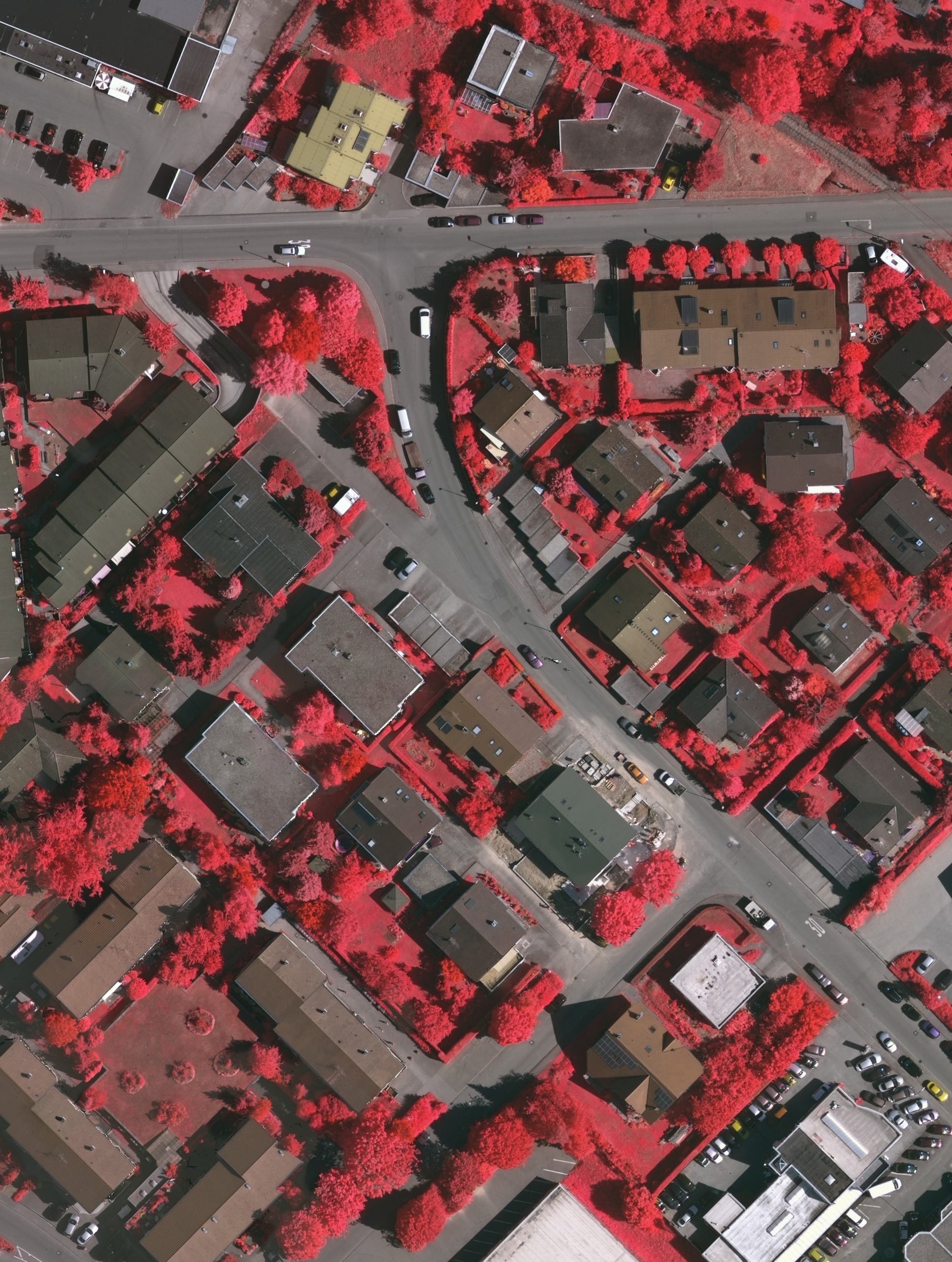}
  \centerline{(a) RG+I image}\medskip
\end{minipage}
\hfill
\begin{minipage}[b]{0.48\linewidth}
  \centering
 \includegraphics[width=1.0\linewidth]{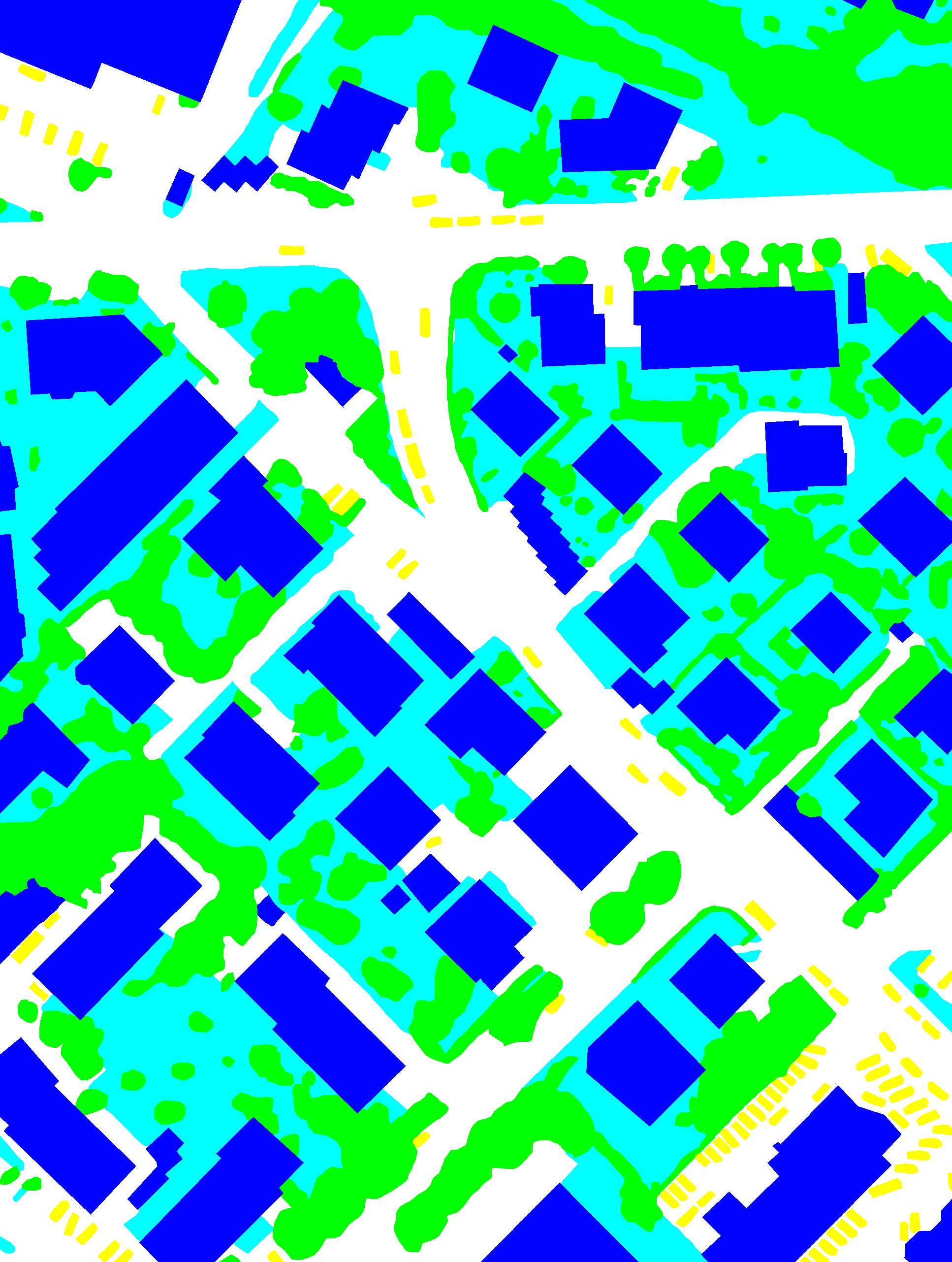}
  \centerline{(b) Ground truth}\medskip
\end{minipage}
\begin{minipage}[b]{.48\linewidth}
  \centering
  \includegraphics[trim=0 0 0 0,clip,width=1.0\linewidth]{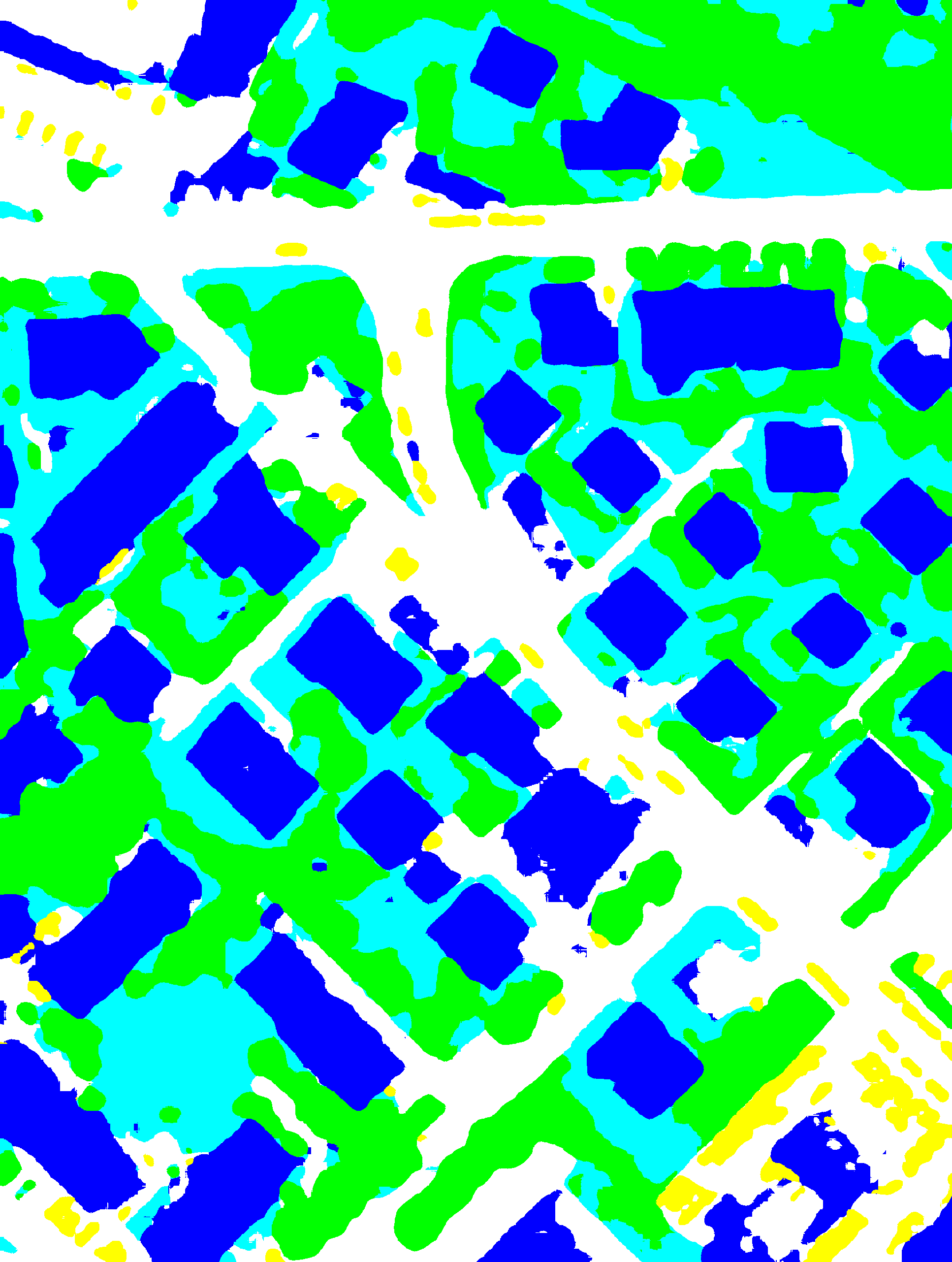}
  \centerline{(c) Segmentation Ensemble}\medskip
\end{minipage}
\hfill
\begin{minipage}[b]{0.48\linewidth}
  \centering
  \includegraphics[trim=0 0 0 0,clip,width=1.0\linewidth]{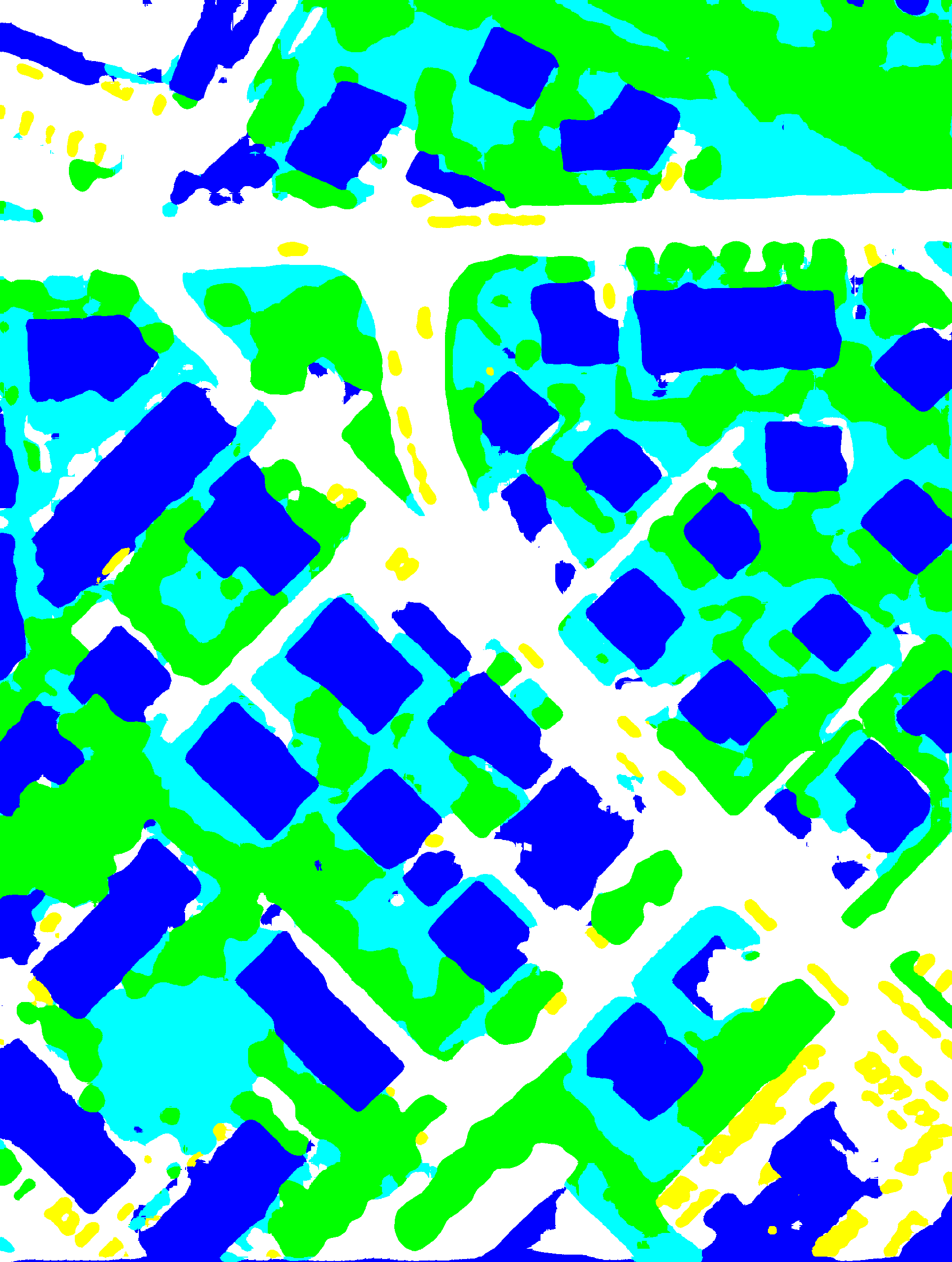}
  \centerline{(d) Segmentation Hallucination}\medskip
\end{minipage}
\caption{Segmentation results for an image in the test dataset.}
\label{fig:res}
\end{figure}

\begin{figure}[htbp]
\begin{minipage}[b]{0.48\linewidth}
  \centering
 \includegraphics[width=1.0\linewidth]{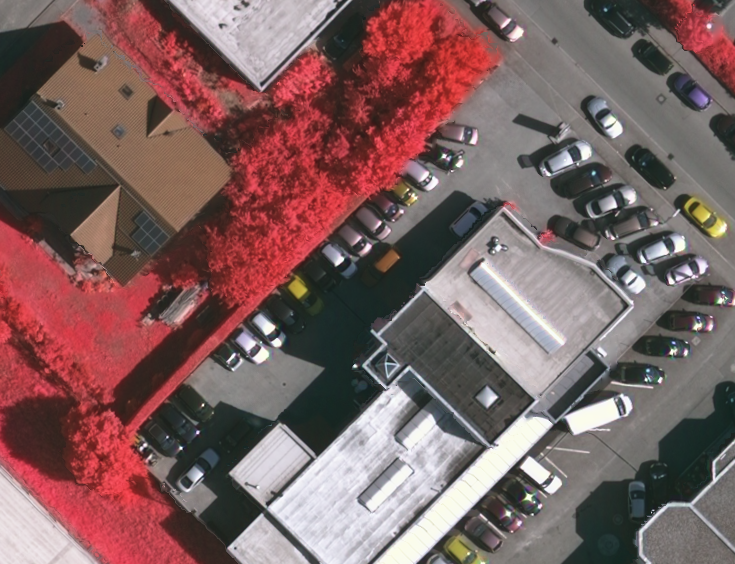}
  \centerline{(a) RG+I image}\medskip
\end{minipage}
\hfill
\begin{minipage}[b]{0.48\linewidth}
  \centering
 \includegraphics[width=1.0\linewidth]{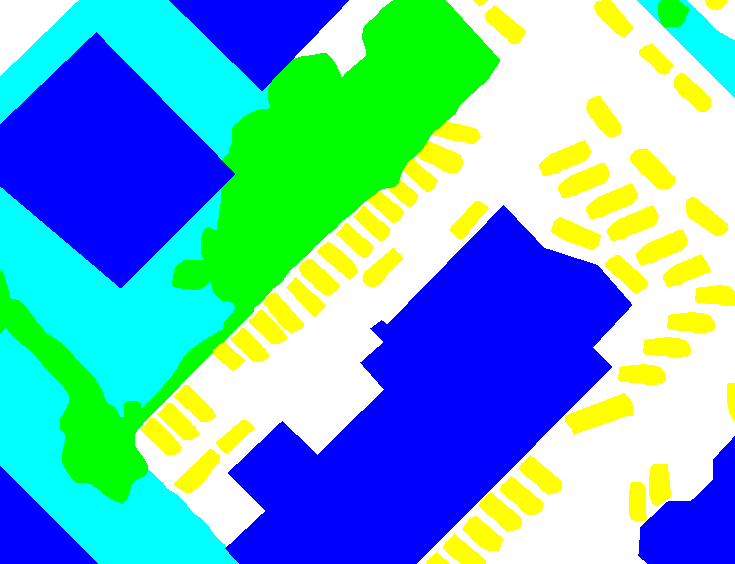}
  \centerline{(b) Ground truth}\medskip
\end{minipage}
\begin{minipage}[b]{0.48\linewidth}
  \centering
 \includegraphics[width=1.0\linewidth]{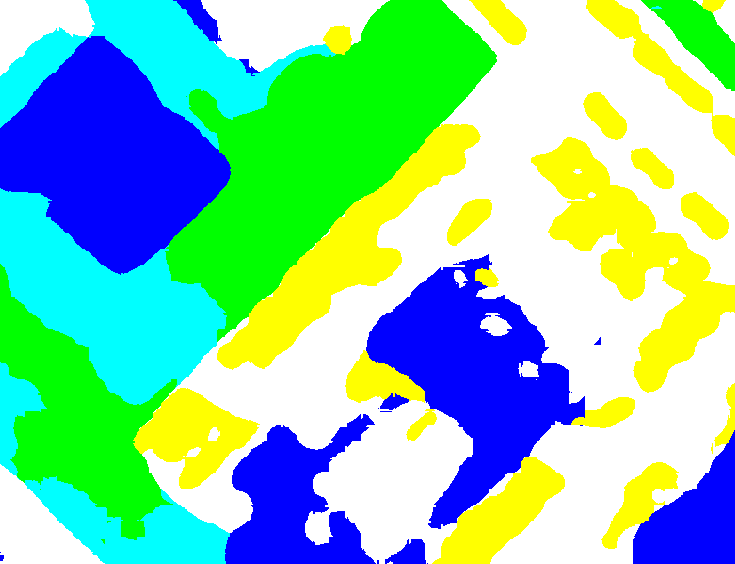}
  \centerline{(c) Ensemble}\medskip
\end{minipage}
\hfill
\begin{minipage}[b]{0.48\linewidth}
  \centering
 \includegraphics[width=1.0\linewidth]{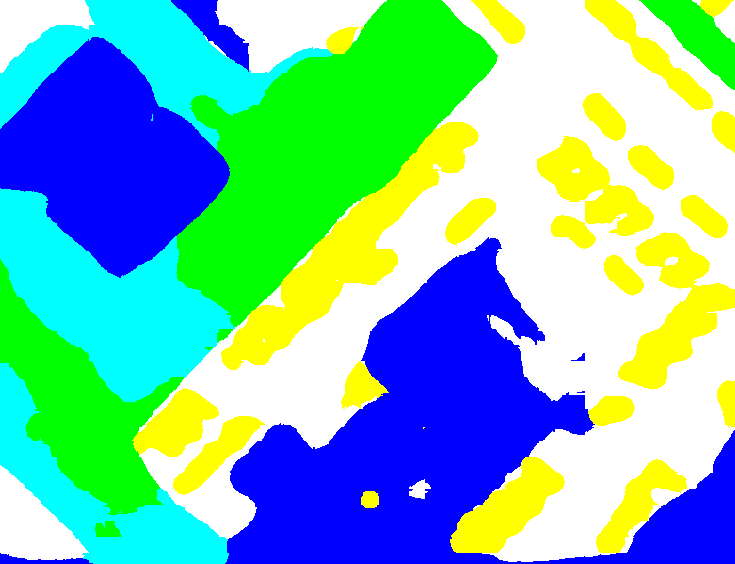}
  \centerline{(d) Hallucination}\medskip
\end{minipage}
\caption{Closeup of the segmentation results for the bottom left corner of the image.}
\label{fig:res_close}
\end{figure}

\begin{figure}[htbp]
\begin{minipage}[b]{0.48\linewidth}
  \centering
 \includegraphics[width=1.0\linewidth]{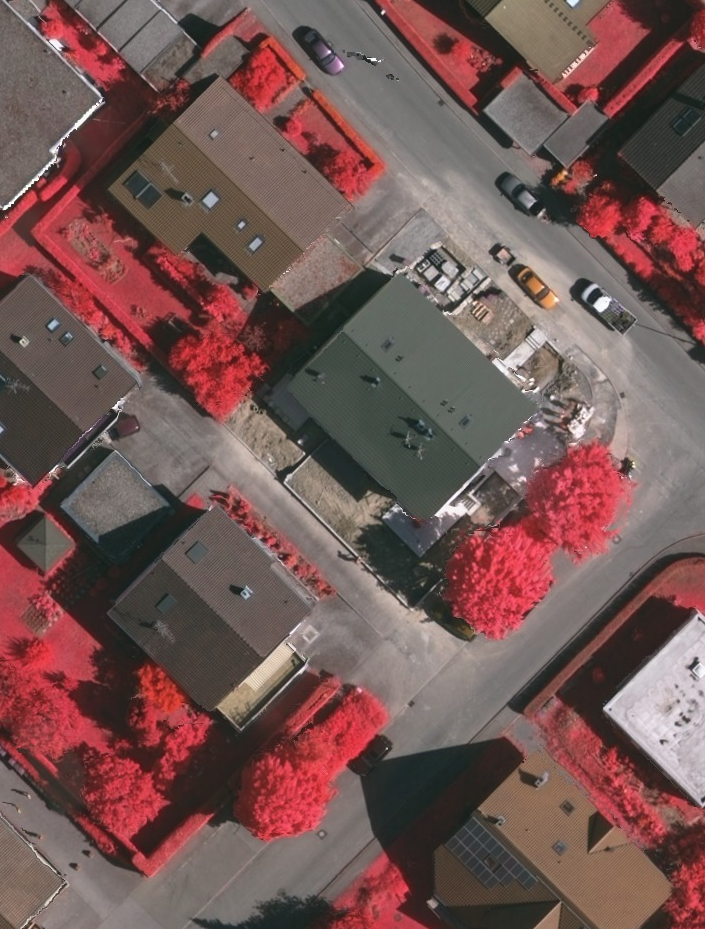}
  \centerline{(a) RG+I image}\medskip
\end{minipage}
\hfill
\begin{minipage}[b]{0.48\linewidth}
  \centering
 \includegraphics[width=1.0\linewidth]{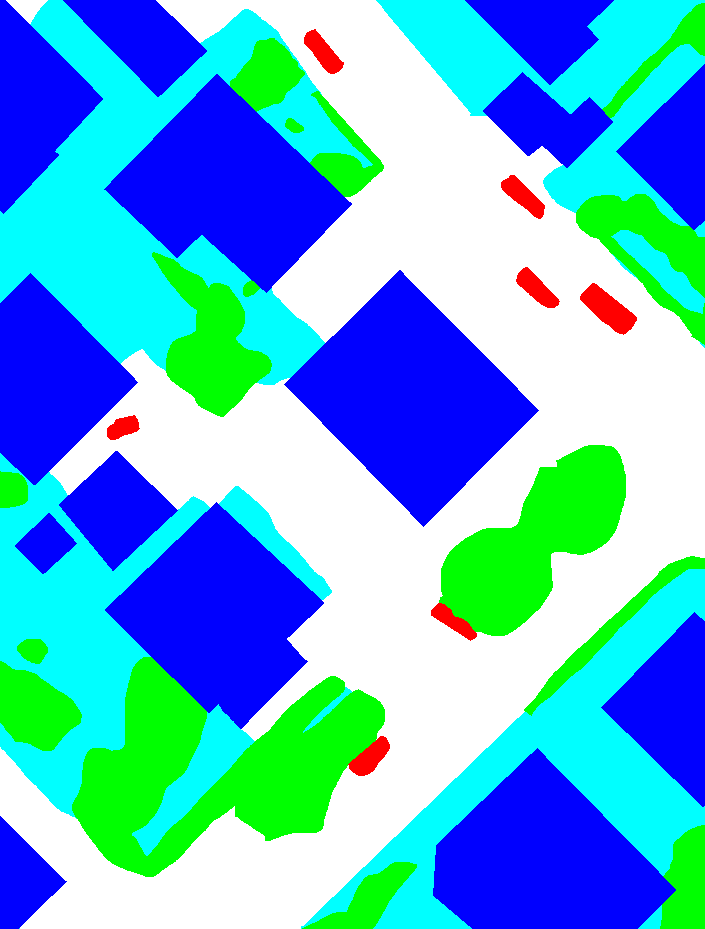}
  \centerline{(b) Ground truth}\medskip
\end{minipage}
\begin{minipage}[b]{0.48\linewidth}
  \centering
 \includegraphics[width=1.0\linewidth]{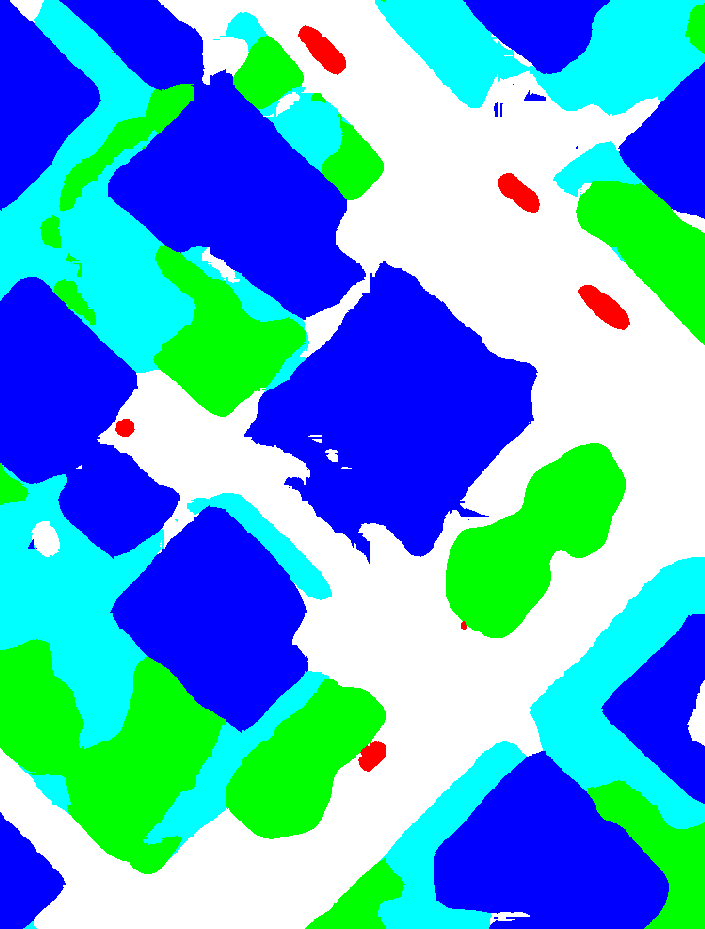}
  \centerline{(c) Hallucination without MFB}\medskip
\end{minipage}
\hfill
\begin{minipage}[b]{0.48\linewidth}
  \centering
 \includegraphics[width=1.0\linewidth]{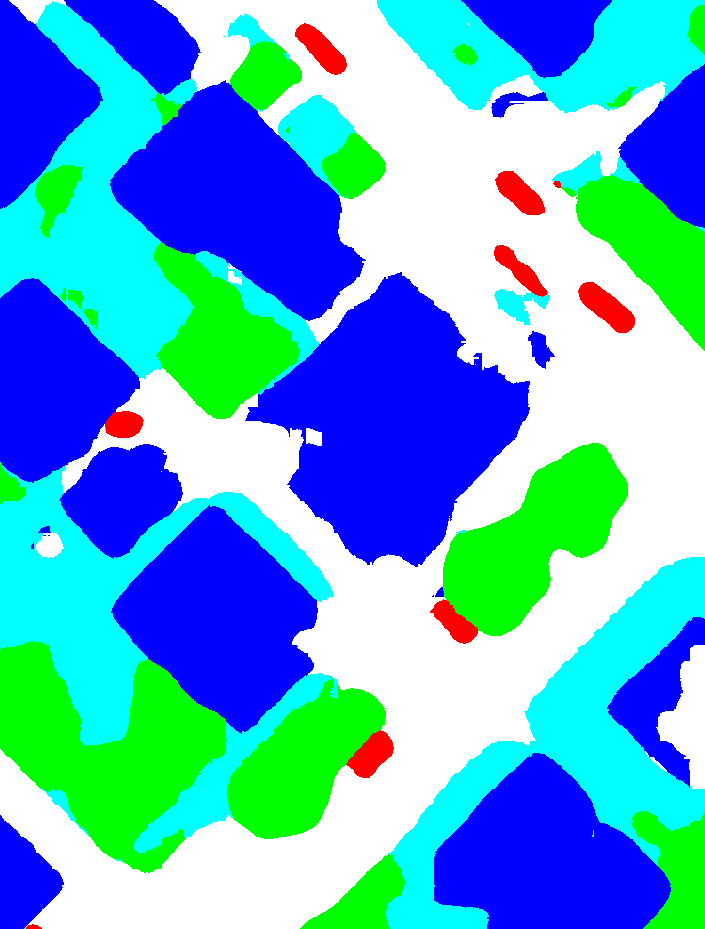}
  \centerline{(d) Hallucination with MFB}\medskip
\end{minipage}
\caption{Illustration of the effect that MFB has on the small car class. To highlight the car class, the color of the car class has been changed to red.}
\label{fig:res_mfb}
\end{figure}

To illustrate some of the differences between the results achieved by the ensemble and the hallucination method, Figure \ref{fig:res} shows one of the RG+I images from the test set, the ground truth, and the achieved segmentation using the proposed method with the median frequency balancing cost function  as well as the RG+I ensemble. It can be seen that both models perform well. However, when comparing the results of the two models it can be observed that the ensemble assigns more impervious surface pixels to the building class. This is as expected, as the color and shape of some roof areas can be considered similar to impervious surfaces. However, including the additional depth information, as done in our proposed method, allows a better separation between these classes as the normalized depth measurements indicate differences between buildings and impervious surfaces. Looking at a close-up of the bottom right corner of the image in Figure \ref{fig:res}, Figure \ref{fig:res_close} illustrates this difference clearly, as large parts of the building gets classified as impervious surface by the RG+I ensemble. By making use of the available depth information during the training phase, the proposed model is able to capture the building class more accurately.

For comparison, we also investigate the performance of a model when the RG+I and depth data are available both during training and testing and we provide the results in Table~\ref{tab:resultsMFBVaihingen}. The overall accuracy for this model is, as expected, higher as more information is available for the network. However, we are able to observe the largest increases in the building and the low vegetation class. This corresponds to our intuition that the depth data is most useful for classes that can clearly be distinguished from similar looking classes with help of a height model. It also illustrates that the hallucination model, with regards to overall accuracy, is able to capture a significant part of the information contained in the depth data.

Table \ref{tab:resultsMFBVaihingen} also illustrates the effect of using medium frequency balancing, as it can be observed that the smaller car class in our experiments generally performs much worse when not using medium frequency balancing. Overall it can be seen that the overall accuracy drops slightly when using the balanced cost function, however, due to the large improvements in the smaller classes, this appears to be a negligible difference. Figure~\ref{fig:res_mfb} shows a qualitative example of the effect of medium frequency balancing. It can be seen that the hallucination network trained without balancing misses some of the cars or only classifies a few of the center pixels in each car. This coincides with our preconception that small classes will overall be neglected in an effort to optimize overall accuracy.

\begin{table*}[t]
\begin{center}
\begin{tabular}{lc|c|c|c|c|c|c|c|c|c|c|c|c|c|c|}
\cline{3-15}
& & \multicolumn{2}{|c|}{Imp Surf} & \multicolumn{2}{c|}{Building} & \multicolumn{2}{c|}{Low veg} & \multicolumn{2}{c|}{Tree} & \multicolumn{2}{c|}{Car} & \multicolumn{3}{c|}{Overall}\\
\cline{1-15}
\multicolumn{1}{|c|}{Method} & \multicolumn{1}{|c|}{MFB} & F1 & Acc & F1 & Acc & F1 & Acc & F1 & Acc & F1 & Acc & Avg F1 & Avg Acc & Acc\\
\cline{1-15} 
\multicolumn{1}{|c|}{RGB} & \cmark & 85.70 & 90.01 & 92.20 & 91.41 & 79.45 & 81.14 & 82.41 & 81.57 & 87.55 & 97.81 & 85.46 & 88.39 & 84.77\\
\cline{2-15}
\multicolumn{1}{|c|}{RGB-ensemble} & \cmark & 86.04 & 89.64 & 92.98 & 92.80 & 81.00 & {\bf83.71} & 83.45 & 82.16 & 89.09 & {\bf98.03} & 86.51 & 89.27 & 85.55\\
\cline{2-15}
\multicolumn{1}{|c|}{Hallucination} & \cmark & {\bf86.77} & {\bf92.16} & {\bf94.03} & {\bf93.10} & {\bf82.39} & 83.03 & {\bf84.72} & {\bf83.41} & {\bf90.98} & 97.95 & {\bf87.78} & {\bf89.93} & {\bf86.56}\\
\hdashline
\multicolumn{1}{|c|}{RGB\&I\&Depth} & \cmark & 88.48 & 92.81 & 96.48 & 96.32 & 80.66 & 82.91 & 83.47 & 81.15 & 88.82 & 97.46 & 87.58 & 90.13 & 87.49\\
\hline
\end{tabular}
\vspace{0.2cm}
\caption{Performance of the different models for the {\bf\scshape Potsdam} dataset when we assume multiple missing modalities, namely infrared and depth. The F1 scores and accuracies are shown as percentages.}
\label{tab:resultsMultiMissing}
\end{center}
\end{table*}

In addition to the F1-measure and the overall accuracy, we also investigated the use of the Intersection over Union (IoU) as an additional evaluation metric. However, as the overall trend is similar to the other metrics, we omitted this evaluation metric in our experimental results. Results for this metric for the experiments reported in this section can be found in the Appendix.

\subsection{Potsdam}
\label{sec:potsdam}
Table \ref{tab:resultsMFBPotsdam} illustrates the results when performing the same set of experiments on the Potsdam dataset. Overall we can observe similar results, where the overall accuracy for the hallucination network outperforms both the RGB+I and the ensemble method. However, the overall increase in accuracy is lower for the Potsdam dataset, which the authors hypothesize is due to the higher resolution, of the images. Due to the higher resolution, there are more factors in the RGB+I bands, which make it possible to distinguish between for example buildings and impervious surfaces.
Table~\ref{tab:resultsMFBPotsdam} also shows that the medium frequency balancing has an effect for the Potsdam dataset. However, also due to the higher resolution, resulting in cars being larger in the small training image patches, the effect is lower compared to the Vaihingen dataset.

Further, we explore the effect of using the trained hallucination model to evaluate images during the test phase where both RGB+I and Depth images are available for the Potsdam dataset (Problem Scenario 2). To illustrate this, we take the trained hallucination model and evaluate its performance when both image modalities are available during testing. The RGB image is fed only into the RGB part of the hallucination network and the depth image is fed into the hallucination networks depth network. The final classification is again produced by applying a softmax to the averaged raw scores of the two networks. 
Note, no additional training is performed and the hallucination model is applied as is.
These results are presented in Table~\ref{tab:resultsHallucinationPotsdam}, which also, to allow the reader to perform an easier comparison, includes a duplicate of the hallucination networks performance when the depth data modality is missing (from Table~\ref{tab:resultsMFBPotsdam}). It can be seen that the hallucination model, when presented with the depth modality also during testing, performs better than when the modality is missing. This means that the trained hallucination model can use the depth modality when it is not missing to achieve better performance, and can, therefore, be used both in situations where both modalities are available, as well as in situations where the depth information is missing (Problem Scenario 2). This flexibility, however, comes at a price, as the overall accuracy for the model when both modalities are available is slightly lower than RGB+I\&Depth in Table~\ref{tab:resultsMFBPotsdam}, where the network is only optimized for situations where both modalities are available.

\subsection{Multiple missing modalities}
\label{sec:missing_experiments}
In this experiment, we evaluate the proposed method on the multiple missing modality scenario, Problem Scenario 3. We assume that the RGB data modality of the Potsdam dataset is available, while the infrared and the depth modalities are missing. Table~\ref{tab:resultsMultiMissing} shows the results for the different methods. It can, like in the case of a single missing modality, be seen that the Hallucination network outperforms both the RGB network and the RGB-ensemble network while performing worse than the network trained on all available modalities. This illustrates the power of the proposed model to not only handle the case of a single missing modality during inference, but also the case of multiple missing data modalities.

\subsection{Experimental setup}
All experiments in this paper were performed using the deep learning framework Caffe~\cite{jia2014caffe} on a single Titan X. Modifications were made to the framework to support the median frequency balancing. Inference on a $1000 \times 1000$ image patch takes 0.08 seconds for the hallucination model illustrated in Figure~\ref{fig:Architecture}. The model was trained for approximately $4$ hours for the Vaihingen dataset and $13$ hours for the Potsdam dataset.
\section{Conclusions and Future Work}
\label{sec:conclusion}
In this paper, we have proposed a method for image segmentation in urban remote sensing that makes use of data modalities that are only available during the training phase. Our experiments show that the method performs better than both a single model using only the available data as well as an ensemble of two models. Additionally, by making use of the medium frequency balancing cost function, we achieve good performances in small classes. We, therefore, consider it an attractive choice for handling missing data modalities in urban remote sensing.
Note that the proposed methodology is applicable not only to urban land cover classification but can be generalized to alternative land cover classification scenarios.

The drawback of the proposed approach for multiple missing data modalities is the fact that it does not scale well with the number of (potentially) missing modalities, as the number of losses grows exponentially. This is in many cases not problematic, as it is common to only consider a limited amount of data modalities. However, in future work we intend to extend this work to a larger amount of missing modalities, trying to devise end-to-end learnable networks that in a scalable way can 'hallucinate' several modalities and that can provide a single model for all combinations of modalities. Also, currently we assume that one of the data modalities is always available, for example in the Vaihingen example we assume that the true ortho photo is available during training and testing. However, in future work, we would like to explore ways of avoiding this restriction. Further, we will investigate transfer learning methodology and the usage of pre-trained models for the sub-networks where pre-trained models are available, such as for example in the case where we have optical image data. Another potential research direction is to utilize larger networks that achieve state-of-the-art performances on segmentation tasks, however, as the focus of this work was to investigate the potential for handling missing data modalities and not achieving overall state-of-the-art performance, this is left for future work.

\section*{Acknowledgment}
We gratefully acknowledge the support of NVIDIA Corporation with the donation of the GPU used for this research and the German Society for Photogrammetry, Remote Sensing and Geoinformation (DGPF) for providing the Vaihingen and the Potsdam dataset \cite{cramer2010dgpf}. This work was partially funded by the Norwegian Research Council FRIPRO grant no.\ 239844 on developing the \emph{Next Generation Learning Machines}.

\ifCLASSOPTIONcaptionsoff
  \newpage
\fi

\bibliographystyle{IEEEtran}
\bibliography{ref.bib}

\begin{table*}[t]
\begin{center}
\begin{tabular}{lc|c|c|c|c|c|c|c|c|c|c|c|c|c|}
\cline{3-14}
 & & \multicolumn{2}{|c|}{Imp Surf} & \multicolumn{2}{c|}{Building} & \multicolumn{2}{c|}{Low veg} & \multicolumn{2}{c|}{Tree} & \multicolumn{2}{c|}{Car} & \multicolumn{2}{c|}{Overall}\\
\cline{1-14}
\multicolumn{1}{|c|}{Method} & \multicolumn{1}{|c|}{MFB} & IoU & Acc & IoU & Acc & IoU & Acc & IoU & Acc & IoU & Acc & Avg IoU & Acc\\
\cline{1-14} 
\multicolumn{1}{|c|}{RG+I} & \cmark & 73.84 & 87.26 & 80.57 & 86.43 & 60.60 & 72.27 & 70.80 & 86.95 & 60.47 & 83.82 & 69.25 & 83.50\\
\cline{2-14}
\multicolumn{1}{|c|}{RG+I-ensemble} & \cmark & 74.38 & 87.83 & 80.71 & 86.14 & 61.45 & 73.52 & 71.35 & 87.01 & 64.14 & {\bf85.19} & 70.40 & 83.86 \\
\cline{2-14}
\multicolumn{1}{|c|}{Hallucination} & \cmark & {\bf77.23} & {\bf89.58} & {\bf83.71} & {\bf88.30} & {\bf62.66} & {\bf74.33} & {\bf72.00} & {\bf87.47} & {\bf67.95} & 84.30 & {\bf72.71} & {\bf85.20} \\
\hdashline
\multicolumn{1}{|c|}{RG+I\&Depth} & \cmark & 79.99 & 89.33 & 87.42 & 91.42 & 62.98 & 75.68 & 72.23 & 87.38 & 69.23 & 83.42 & 74.37 & 86.33 \\
\hline
\hline
\multicolumn{1}{|c|}{RG+I} & \xmark & 76.73 & 89.13 & 84.31 & 89.81 & 60.38 & 75.18 & 74.93 & 86.36 & 41.85 & 44.38 & 67.64 & 84.96\\
\hline
\multicolumn{1}{|c|}{RG+I-ensemble} & \xmark & 77.21 & 89.47 & 84.58 & 89.79 & 60.92 & {\bf75.74} & 75.07 & 86.39 & 42.87 & 45.09 & 68.13 & 85.17 \\
\hline
\multicolumn{1}{|c|}{Hallucination} & \xmark & {\bf78.88} & {\bf91.14} & {\bf86.04} & {\bf90.58} & {\bf61.99} & 74.31 & {\bf76.19} & {\bf88.34} & {\bf59.68} & {\bf62.90} & {\bf72.56} & {\bf86.22}\\
\hdashline
\multicolumn{1}{|c|}{RG+I\&Depth} & \xmark & 78.84 & 91.35 & 87.02 & 91.36 & 62.50 & 75.69 & 76.51 & 87.54 & 48.49 & 50.01 & 70.67 & 86.39\\
\hline
\end{tabular}
\vspace{0.2cm}
\caption{Performance of the different models for the {\bf\scshape Vaihingen} dataset. The IoU scores and accuracies are shown as percentages. Bold numbers indicate the best accuracy among the first three models. The final model $\text{RG+I\&Depth}$ is used as a reference to illustrate the overall accuracy that could be achieved by a model if all data modalities are available and no hallucination network is employed.}
\label{tab:resultsMFBVaihingenIOU}
\end{center}
\end{table*}

\appendix
\label{sec:appendix}
Table~\ref{tab:resultsMFBVaihingenIOU} illustrates another performance metric for the Results in Table~\ref{tab:resultsMFBVaihingen}, namely, the Intersection over Union (IoU). Following the notation in Section~\ref{sec:dataset}, the IoU is defined as $n_{ii}/(t_i + \sum_j n_{ji} - n_{ii})$. $n_{ij}$ denotes the number of pixels of class $i$ that are classified as class $j$ and $t_i$ the total number of pixels in class $i$. Overall, we observed that the trend for the IoU was well aligned with the F1-measures and has therefore, due to reasons of brevity, been omitted for the remaining tables.

\begin{IEEEbiography}[{\includegraphics[width=1in,height=1.25in,clip,keepaspectratio]{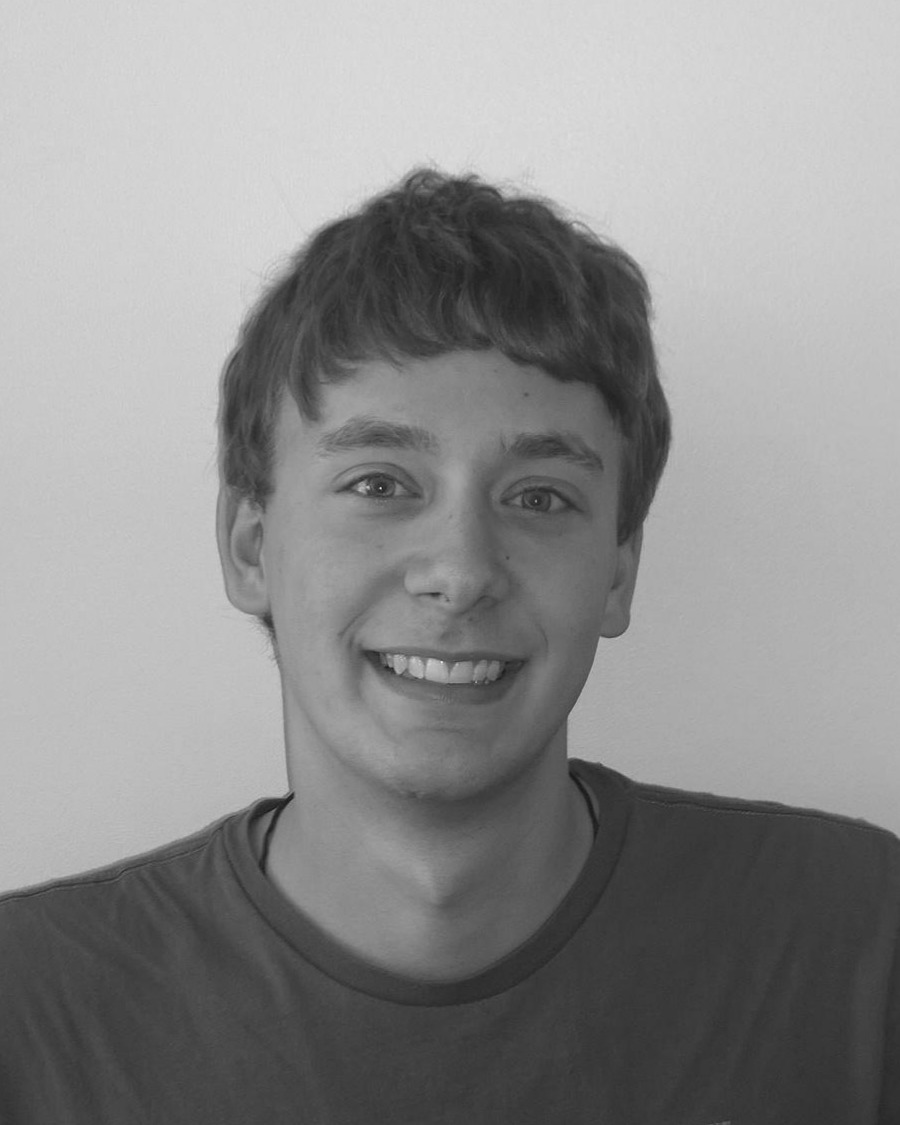}}]{Michael Kampffmeyer}
received Master’s degrees at the Physics and Technology (2014) and Computer Science Department (2015) at UiT The Arctic University of Norway, Troms{\o}, Norway, where he currently pursues a PhD in Machine Learning. His research interests include the development of unsupervised deep learning methods for representation learning and clustering by utilizing ideas from kernel machines and information theoretic learning. Further, he is interested in computer vision, especially related to remote sensing and health applications. His paper 'Deep Kernelized Autoencoders' with S. L{\o}kse, F. M. Bianchi, R. Jenssen and L. Livi won the Best Student Paper Award at the Scandinavian Conference on Image Analysis, 2017. Since September 2017, he has been a Guest Researcher in the lab of Eric P. Xing at Carnegie Mellon University, Pittsburgh, PA, USA, where he will stay until July 2018. For more details visit https://sites.google.com/view/michaelkampffmeyer/.
\end{IEEEbiography}
\enlargethispage{-2.8in}
\vspace{-1.6cm}
\begin{IEEEbiography}[{\includegraphics[width=1in,height=1.25in,clip,keepaspectratio]{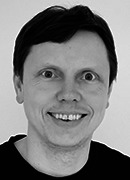}}]{Arnt-B{\o}rre~Salberg}
(M’03) received the Diploma in applied physics and the Ph.D. degree in physics from the University of Troms{\o}, Troms{\o}, Norway, in 1998 and 2003, respectively.
From August 2001 to June 2002, he was a Visiting Researcher with the U.S. Army Research Laboratory, Adelphi, MD, USA. From February 2003 to December 2005, he had a postdoctoral and research
position with the Institute of Marine Research, Troms{\o}, Norway. From December 2005 to October 2008, he was the Head of R\&D with Dolphiscan AS, Moelv, Norway. Since October 2008, he has been with the Norwegian Computing Center, Oslo, Norway, where he is currently a Senior Research Scientist in earth observation. His research interests are in the areas of earth observation, signal and image processing, machine learning, and computer vision.
\end{IEEEbiography}
\begin{IEEEbiography}[{\includegraphics[width=1in,height=1.25in,clip,keepaspectratio]{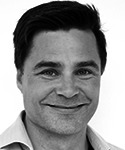}}]{Robert~Jenssen}
(M’02) received the PhD (Dr. Scient.) in Electrical Engineering from the University of Troms{\o}, in 2005.
Currently, he is an Associate Professor with the Department of Physics and Technology, UiT The Arctic University of Norway, Troms{\o}, Norway. He directs the UiT Machine Learning Group: http://site.uit.no/ml. The group's focus is on innovating information theoretic learning, kernel machines, and deep learning research, an activity that is funded, e.g., over the Norwegian Research Council's prestigious FRIPRO program, grant 239844 (Next-Generation Learning Machines). He is also a Research Professor with the Norwegian Computing Center, Oslo, Norway. He was a Guest Researcher with the Technical University of Denmark, Kongens Lyngby, Denmark, from 2012 to 2013, with the Technical University of Berlin, Berlin, Germany, from 2008 to 2009, and with the University of Florida, Gainesville, FL, USA, from 2002 to 2003, spring 2004, and spring 2018. Jenssen was a senior researcher at the Norwegian Center for E-Health Research, 2012-2015. Jenssen received the Honorable Mention for the 2003 Pattern Recognition Journal Best Paper Award, the 2005 IEEE ICASSP Outstanding Student Paper Award, and the 2007 UiT Young Investigator Award. His paper “Kernel Entropy Component Analysis” was the featured paper of the May 2010 issue of the IEEE Transactions on Pattern Analysis and Machine Intelligence. In addition, his paper “Kernel Entropy Component Analysis for Remote Sensing Image Clustering” with L. Gomez-Chova and G. Camps-Valls received the IEEE Geoscience and Remote Sensing Society Letters Prize Paper Award in 2013. His student M. Kampffmeyer won the Best Student Paper Award for the paper 'Deep Kernelized Autoencoders' at the Scandinavian Conference on Image Analysis, 2017. Jenssen is on the IEEE Technical Committee on Machine Learning for Signal Processing (MLSP), he is the president of the Norwegian section of IAPR (NOBIM - nobim.no) and serves on the IAPR Governing Board. He is an associate editor with the journal Pattern Recognition since 2010. 
\end{IEEEbiography}

\end{document}